%% file: neurips_2026.tex
\documentclass{article}

\PassOptionsToPackage{numbers}{natbib}

\usepackage[main, final]{neurips_2026}

\usepackage[utf8]{inputenc} % allow utf-8 input
\usepackage[T1]{fontenc}    % use 8-bit T1 fonts
\usepackage{hyperref}       % hyperlinks
\usepackage{url}            % simple URL typesetting
\usepackage{booktabs}       % professional-quality tables
\usepackage{amsfonts}       % blackboard math symbols
\usepackage{nicefrac}       % compact symbols for 1/2, etc.
\usepackage{microtype}      % microtypography
\usepackage{xcolor}         % colors
\usepackage{wrapfig}

\usepackage{algorithm}
\usepackage{algpseudocode}
\usepackage{adjustbox}
\usepackage{multirow}
\usepackage{geometry}
\usepackage{caption} 
\usepackage{amsmath}
\usepackage{subcaption}
\usepackage{cleveref}

\title{Parameterized Stripe Attention for Efficient Video Generation}

\author{%
  Xingyu Jia, Baole Ai, Ang Wang, Kang Zhao, Yong Li \\
  Alibaba Group
}

\begin{document}

\maketitle

\begin{abstract}

    Diffusion Transformers (DiTs) enable high-quality video generation but suffer from substantial inference latency, primarily attributable to the computationally expensive full spatio-temporal attention. While sparse attention methods offer potential solutions, existing approaches face an inherent flexibility--efficiency dilemma: predefined masks lack the flexibility to capture diverse attention patterns, while runtime-determined masks introduce overheads and sacrifice hardware efficiency. 
    We identify the lack of a unified structural characterization of DiT attention as a key limitation of existing methods, 
    and establish that video DiT attention exhibits \textbf{periodic diagonal stripe structures} along both temporal and spatial dimensions. 
    To formally encode these structured patterns within a single efficient kernel,
    we present {\bf PSA}, a parameterized stripe attention that formalizes the observed stripe regularity, unifying diverse attention patterns for efficient mask generation. This unified representation enables a single hardware-efficient CUDA kernel to process all sparse patterns, achieving FlashAttention-3-level Model FLOPs Utilization. To determine optimal sparsity configurations, we propose a training-free offline search algorithm that automatically maximizes sparsity under a specified error tolerance for each attention head. Experiments on HunyuanVideo and Wan~2.1 demonstrate that PSA achieves 1.57$\times$ and 1.37$\times$ end-to-end speedups over FlashAttention-3 baselines, with acceptable visual quality degradation.

\end{abstract}

\input{sec/1_intro}
\input{sec/2_related_work}

\input{sec/3_method}
\input{sec/4_experiment}

\input{sec/5_conclusion}

\input{sec/acknowledgements}

\bibliographystyle{splncs04}
\bibliography{main}

%%%%%%%%%%%%%%%%%%%%%%%%%%%%%%%%%%%%%%%%%%%%%%%%%%%%%%%%%%%%

\appendix
\input{sec/X_suppl}

% \section{Technical appendices and supplementary material}
% Technical appendices with additional results, figures, graphs, and proofs may be submitted with the paper submission before the full submission deadline (see above). You can upload a ZIP file for videos or code, but do not upload a separate PDF file for the appendix. There is no page limit for the technical appendices. 

% Note: Think of the appendix as ``optional reading'' for reviewers. The paper must be able to stand alone without the appendix; for example, adding critical experiments that support the main claims to an appendix is inappropriate. 

%%%%%%%%%%%%%%%%%%%%%%%%%%%%%%%%%%%%%%%%%%%%%%%%%%%%%%%%%%%%
\clearpage
\newpage
\input{checklist.tex}

\end{document}

%% file: sec/1_intro.tex
\section{Introduction}
\label{sec:intro}

Diffusion Transformers (DiTs)~\cite{peebles2023scalable} have rapidly emerged as a foundational architecture for image and video generation. 
DiTs employ Transformers~\cite{vaswani2017attention} as the backbone and leverage algorithms such as Diffusion~\cite{ddim,lu2022dpm,lu2025dpm++,peebles2023scalable} or Flow Matching~\cite{lipman2022flowmatching,esser2024rectifiedflow} to iteratively generate images and videos from noise. Prominent open-source models such as Wan~2.1~\cite{wan2025wan} and HunyuanVideo~\cite{kong2024hunyuanvideo} leverage the DiT architecture for tasks like text-to-video and image-to-video synthesis, producing high-quality, highly dynamic video content.
However, DiT inference requires multi-step denoising with enormous computational costs and prolonged latency. 
For instance, generating a 5-second, 720p video with Wan~2.1 on an NVIDIA H100 GPU takes nearly 30 minutes~\cite{wan2025wan}, severely limiting practical applications. Since these models utilize full spatio-temporal 3D attention~\cite{yang2024cogvideox}, the attention computation constitutes the primary bottleneck, accounting for over 60\%~\cite{kong2024hunyuanvideo} of the total inference time. Consequently, designing more efficient attention mechanisms is a critical optimization target. Sparse attention---selectively computing only the most informative token interactions---offers a promising avenue, provided the attention distribution exhibits exploitable structure.

% \begin{figure}[t!]
%   \centering
%   \begin{minipage}[b]{0.355\linewidth}
%     \centering
%     \includegraphics[width=\linewidth]{pic/sec1/heat-map.pdf}
%     \captionof{figure}{Attention weights from different heads in Wan~2.1 and HunyuanVideo.}
%     \label{fig:heatmap}
%   \end{minipage}
%   \hfill
%   \begin{minipage}[b]{0.545\linewidth}
%     \centering
%     \includegraphics[width=\linewidth]{pic/sec3/attn_patterns.pdf}
%     \captionof{figure}{Visualization of the \texttt{PSA} parameterization that captures various attention sparse patterns.}
%     \label{fig:param_vis}
%   \end{minipage}
% \end{figure}

\begin{figure}[t!]
  \centering
  \begin{minipage}[t]{0.355\linewidth}
    \centering
    \adjustbox{height=5cm}{
      \includegraphics[width=\linewidth]{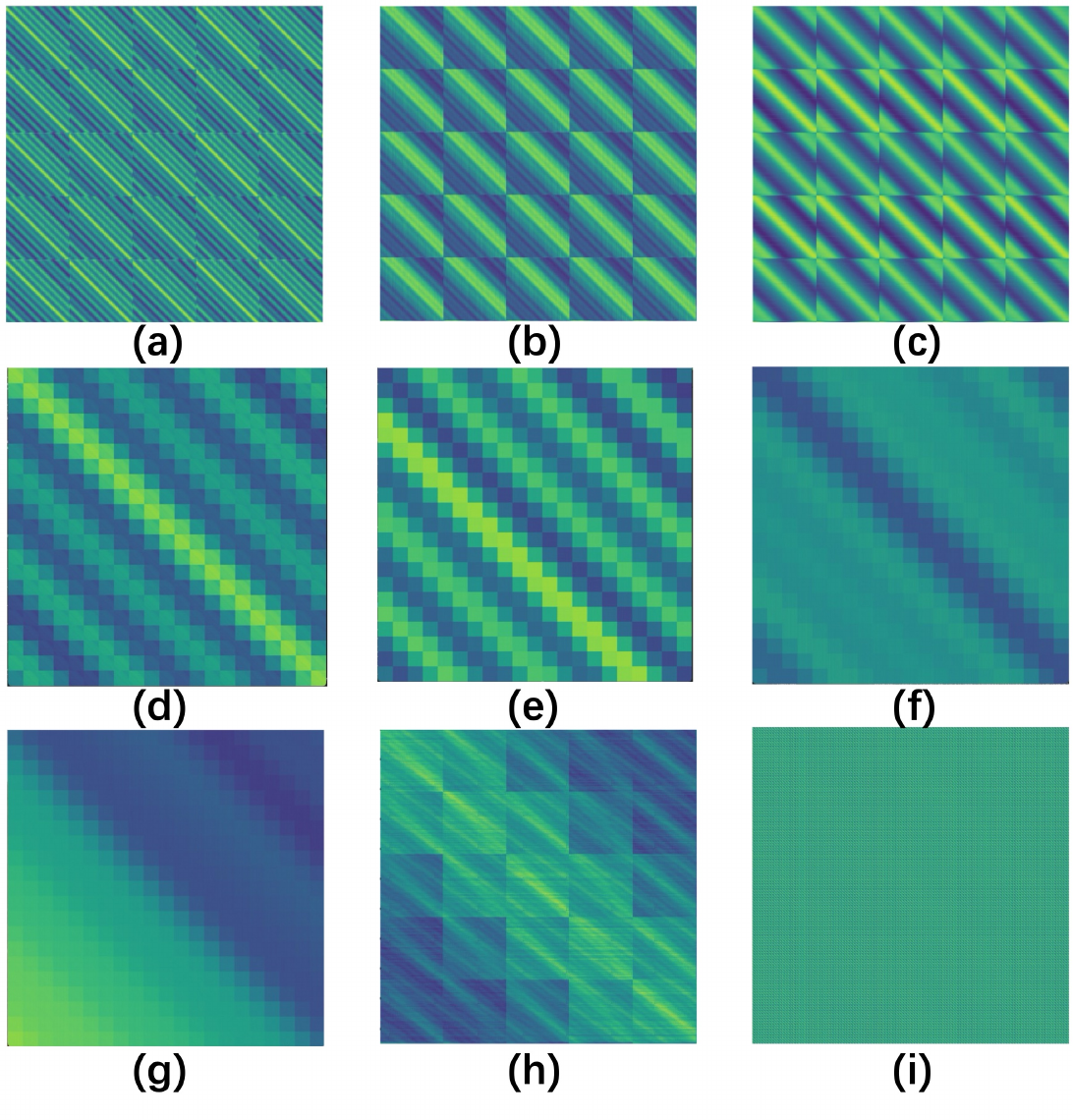}
    }
    \captionof{figure}{Heat maps in Wan~2.1 and HunyuanVideo.}
    \label{fig:heatmap}
  \end{minipage}
  \hfill
  \begin{minipage}[t]{0.55\linewidth}
    \centering
    \adjustbox{height=5cm}{
      \includegraphics[width=\linewidth]{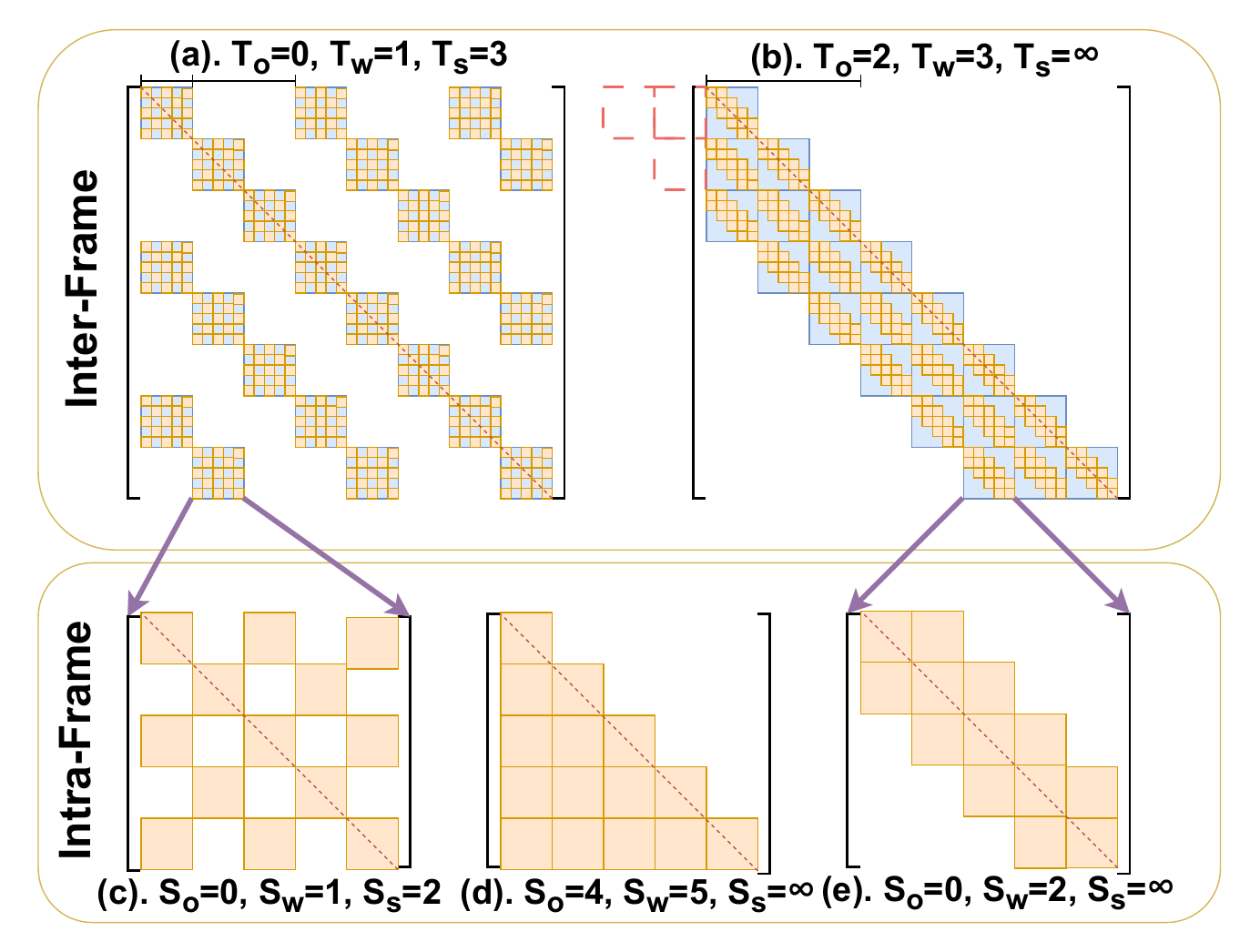}
    }
    \captionof{figure}{Visualization of the \texttt{PSA} parameterization that captures various attention sparse patterns.}
    \label{fig:param_vis}
  \end{minipage}
\end{figure}

Existing sparse attention methods for DiTs lack a principled structural characterization of attention distributions, leading to an inherent trade-off between flexibility and efficiency.
The first category relies on \textbf{predefined} masks designed from empirical observations rather than principled analysis.
SVG~\cite{svg} identifies only two fixed patterns—temporal and spatial heads—without analyzing why only these two arise or where the observed periodicity originates; consequently, it cannot capture the diverse stride, offset, and hybrid patterns revealed by our analysis.
STA~\cite{sta} achieves efficient local computation via tile-wise sliding windows, but assumes only local attention and cannot represent long-range, strided, or hybrid structures as shown in~\cref{fig:heatmap}.
Because these methods are not grounded in a comprehensive characterization of attention patterns, their fixed masks inevitably fail to cover the full diversity of sparse structures.
The second category pursues more flexible pattern representations by determining the sparse mask dynamically \textbf{at runtime}, effectively treating attention distributions as arbitrary. For example,
SVG2~\cite{svg2} clusters Q and K tokens separately to enable flexible sparse patterns, overlooking the structural regularity revealed by our analysis.
This abandonment of structure comes at a cost: the irregular cluster-based kernel's reliance on FlashInfer~\cite{ye2025flashinferefficientcustomizableattention} precludes Tensor Memory Accelerator (TMA) utilization and introduces non-trivial padding overhead ($\sim$20\% at 70\% sparsity) despite token permutation; and the clustering procedure itself incurs additional computational cost ($\sim$10\% at 70\% sparsity).
Consequently, the flexibility–efficiency trade-off in existing methods stems from the lack of a principled, unified framework to characterize sparse attention: methods 
with fixed masks fail to cover the full diversity of DiT-specific sparse 
structures, while methods pursuing flexibility sacrifice efficiency 
by treating attention distributions as arbitrary rather than structured.
This motivates the need for a more mechanistic understanding of DiT attention patterns that can support both broad pattern coverage and high-performance implementation.

Concretely, sparse attention for video DiTs involves three main challenges:
% 1) A complete theoretical analysis and characterization of the sparse structures in attention maps is needed. 
(1) A principled characterization of the DiT attention structure space is 
needed, providing a unified and sufficiently expressive description of the diverse sparse patterns 
that emerge across heads, layers, and denoising timesteps. 
(2) A hardware-aware implementation must be developed that supports various mask configurations while maximizing GPU utilization. 
(3) An efficient search method is required to determine per-head sparse configurations without incurring additional inference-time overhead.

To address these challenges, we propose \texttt{PSA}, a parameterized stripe sparse attention framework.
First, we reveal that video DiT attention exhibits \textbf{periodic diagonal stripe structures} along both temporal and spatial dimensions, through theoretical derivation (via RoPE-3D logit decomposition and video spatio-temporal redundancy) and large-scale empirical validation across all attention heads in HunyuanVideo and Wan~2.1 (\cref{fig:heatmap}), identifying four primary patterns: intra-frame, inter-frame, hybrid, and uniform (see \cref{sec:unified_pattern_para} for details).
Second, based on this structural understanding, we design a parameterized mask $\mathcal{M}(T_w, T_o, T_s, S_w, S_o, S_s)$ that provides a unified representation of periodic stripe sparse patterns across attention heads, with quantitative evaluation demonstrating high-fidelity approximation of full attention outputs (average cosine similarity of 0.9831).
Leveraging this unified representation, we support different patterns with a single kernel, eliminating multi-kernel launch overhead.
Our implementation adopts block-wise partitioning as the fundamental granularity to fully exploit modern GPU hardware resources, achieving 61\% Model FLOPs Utilization (MFU) on HunyuanVideo—close to FlashAttention-3 levels—thereby resolving the flexibility--efficiency tension.
Finally, observing strong similarity in attention masks across different prompts, we design an offline parallel search algorithm that requires no training and introduces no runtime search overhead,
assigning the mask with the highest sparsity to each head while maintaining error bounds.
Our parametric pattern representation and offline search achieve both objectives: head-adapted patterns are determined prior to inference, thereby \textbf{eliminating runtime overhead while preserving adaptability}.
We validate our method on HunyuanVideo and Wan~2.1, achieving 1.57$\times$ and 1.37$\times$ end-to-end speedups over FlashAttention-3, respectively, with acceptable quality degradation (PSNR $>$ 23).
In summary, our main contributions are as follows.
\begin{enumerate}
\setlength{\itemsep}{1pt}
\setlength{\parsep}{0pt}
\setlength{\parskip}{0pt}
\item We reveal that video DiT attention exhibits periodic diagonal stripe structures along both temporal and spatial dimensions, and provide both theoretical derivation and large-scale empirical validation to establish this structural property.
\item We propose a parameterized mask $\mathcal{M}(T_w, T_o, T_s, S_w, S_o, S_s)$ that unifies diverse sparse patterns, enabling a single hardware-efficient CUDA kernel to process all supported patterns with FlashAttention-3-level MFU, resolving the flexibility--efficiency tension. We further introduce 3D-padding to support arbitrary video resolutions with minimal overhead ($\approx$5\% for 720p).
\item We develop a training-free offline search algorithm that maximizes sparsity under a constrained error threshold for each attention head, with support for distributed parallel search to accelerate the process.
\item The \texttt{PSA} achieves substantial acceleration without compromising effectiveness, outperforming state-of-the-art methods such as SVG2 and STA in both accuracy and efficiency.
\end{enumerate}

%% file: sec/2_related_work.tex
\section{Related Work}
\label{sec:related_work}

Attention mechanisms are pivotal to the success of Transformers~\cite{vaswani2017attention}, 
yet their quadratic complexity poses a fundamental efficiency bottleneck. Existing efficient 
attention methods for video generation primarily construct sparse masks by exploiting the 
distributional characteristics of attention heatmaps, falling into two paradigms.
The first paradigm relies on \textbf{predefined sparse mask formats}. 
SVG~\cite{svg} classifies attention heads into spatial and temporal types and employs 
online profiling to generate sparse masks. 
VORTA~\cite{sun2025vorta} trains a lightweight router to dynamically select among full 
attention, sliding window, and coreset attention patterns. 
Sparse-vDiT~\cite{sparse_vdit} selects from a set of predefined structural patterns, 
including diagonal, multi-diagonal, and vertical-stripe sparsity. 
Additionally, other window-based attention methods exploit the locality of attention. NATTEN~\cite{hassani2023neighborhood} implements sliding window attention for vision tasks, localizing self-attention to nearest neighborhoods.
STA~\cite{sta} employs tile-wise window attention for improved memory access and parallelism.
The second paradigm \textbf{determines sparse masks dynamically} during inference based on attention scores.
VMoBA~\cite{wu2025vmoba} extends Moba~\cite{lu2025moba} to adapt spatio-temporal attention within DiTs.
MOD-DiT~\cite{liu2026mixturedistributionsmattersdynamic} avoids the sampling overhead of runtime mask determination by linearly extrapolating pattern intensities from early denoising steps.
RoPeSLR~\cite{liu2026ropeslr3dropedrivensparselowrank} combines block-sparse attention with a trained low-rank compensator to restore dropped background contributions.
SVG2~\cite{svg2} improves efficiency through $k$-means clustering of $Q$ and $K$ 
vectors, followed by token permutation and top-$k$ cluster selection to identify 
the most relevant attention regions.

However, existing methods suffer from notable limitations: predefined pattern approaches 
are constrained by a limited pattern set, while dynamic mask generation leads to non-trivial overhead and poor efficiency, yielding limited speedup over optimized dense implementations such as 
FlashAttention~\cite{dao2023flashattention2,shah2024flashattention}. 
In contrast, \texttt{PSA} uniformly parameterizes diverse sparse patterns with 
hardware-aware kernels, achieving superior flexibility and efficiency.

% V2 end

% Additionally, other window-based attention methods exploit the locality of attention. NATTEN~\cite{hassani2023neighborhood} implements a Sliding Window Attention (SWA) mechanism for vision tasks, localizing self-attention for each pixel to its nearest neighborhood. STA~\cite{sta} employs a tile-wise window attention implementation (as opposed to token-wise SWA) to enhance memory access efficiency and parallelism. We demonstrate that STA can be viewed as a specific instance within our proposed sparse representation framework.

%% file: sec/3_method.tex
\section{Method}
\label{sec:method}

% \begin{figure}[t!]
%   \centering
%   \includegraphics[width=1\linewidth]{pic/sec3/Method-main.pdf}
  
%   \caption{An overview of the \texttt{PSA} framework. 
%     \textbf{(a) Unified Representation:} The parametric model $\mathcal{M}$ captures diverse per-head sparse attention patterns, covering both intra-frame and inter-frame structures with varying widths, offsets, and strides. 
%     \textbf{(b) Efficient Kernel:} A dedicated CUDA kernel enables high-performance, hardware-aware computation across all supported sparse patterns. 
%     \textbf{(c) Automatic Search:} \texttt{PSA-Search} automatically identifies the optimal sparse configuration for each head by maximizing sparsity subject to a predefined output degradation threshold.}

%   \label{fig:method-main}
% \end{figure}

%-------------------------------------------------------------------------
\subsection{Structural Analysis of Attention Patterns}
\label{sec:unified_pattern_para}

We begin by formally characterizing the structural properties of attention maps in video DiTs, aiming to move beyond empirical observation toward a mechanistic understanding of the patterns governing these distributions. Our analysis reveals that attention in video DiTs is neither arbitrary nor purely local; rather, it consistently manifests as periodic diagonal stripe structures, parameterized by their widths, offsets, and strides.
We verify this structural analysis through a comprehensive examination of attention maps in HunyuanVideo and Wan~2.1, identifying four primary attention patterns as shown in \cref{fig:heatmap}:
\begin{enumerate}
\setlength{\itemsep}{1pt}
\setlength{\parsep}{0pt}
\setlength{\parskip}{0pt}
    \item \textbf{Intra-frame} [Fig.~(a)--(c)]: diagonal stripes with
    varying widths, offsets, and strides within each frame, repeating
    identically across frames.

    \item \textbf{Inter-frame} [Fig.~(d)--(g)]: relatively uniform
    intra-frame weights, with diagonal band structures of varying widths, offsets, and strides emerging between frames.

    \item \textbf{Hybrid} [Fig.~(h)]: a combination of intra-frame and inter-frame patterns.

    \item \textbf{Uniform} [Fig.~(i)]: attention weights uniformly
    distributed across all positions.
\end{enumerate}

We show that this structure arises from two complementary properties: \textbf{(1) video spatio-temporal redundancy}, which concentrates high attention scores within local neighborhoods, and \textbf{(2) RoPE-3D cosine periodicity}, which causes attention scores to repeat at axis-dependent intervals.
% We now establish each property and their joint effect.

\noindent
\textbf{Property 1: Video spatio-temporal redundancy.}
Videos exhibit rich spatio-temporal redundancy at multiple scales~\cite{pan2021vared2videoadaptiveredundancy, wang2025retakereducingtemporalknowledge}.
Since attention is the only component in a DiT block that computes pairwise token interactions, this redundancy is directly reflected in the attention score matrix $\mathbf{A}$.
Concretely, fixing any two axes and varying the third, high attention scores concentrate
within locality radii $\varepsilon_t$, $\varepsilon_h$, $\varepsilon_w$ along the temporal,
height, and width axes, respectively:
\begin{equation}
    |\Delta t|<\varepsilon_t \;\;(h,w\text{ fixed}),\quad
    |\Delta h|<\varepsilon_h \;\;(t,w\text{ fixed}),\quad
    |\Delta w|<\varepsilon_w \;\;(t,h\text{ fixed}),
    \label{eq:locality}
\end{equation}
where $\Delta t = t - t'$, $\Delta h = h - h'$, $\Delta w = w - w'$ denote relative
offsets along each axis.
% When $\mathbf{A}$ is visualized as a 2D heatmap, this locality means that high attention
% weights concentrate near the main diagonal---forming the \textbf{width} of the observed stripes.
% The width of inter-frame stripes is governed by $\varepsilon_t$, while the width of intra-frame stripes is governed by $\varepsilon_w$.
In the 2D heatmap, the locality conditions in \cref{eq:locality} concentrate high attention along diagonal bands---forming the stripe \textbf{width} ($\varepsilon_t$ inter-frame, $\varepsilon_w$ intra-frame)---while column-wise locality ($\varepsilon_h$) produces multiple parallel diagonals with layout-determined stride (see Appendix~\ref{sec:locality_stripes}).

\noindent
\textbf{Property 2: RoPE-3D cosine periodicity.}
% Video redundancy alone, however, cannot explain why stripes \emph{repeat} at fixed intervals
% (\textbf{stride}) or why they may shift away from the main diagonal (\textbf{offset}).
% We now show that the rotary position encoding in video DiTs introduces precisely this periodicity.
% Rotary Position Embedding (RoPE)~\cite{su2024roformer} has become the de facto positional encoding for attention mechanisms in transformers.
% Video DiTs apply RoPE-3D~\cite{kong2024hunyuanvideo, wan2025wan}, which partitions the
% embedding channels into three disjoint subsets $M_t$, $M_h$, and $M_w$, encoding temporal,
% height, and width positions, respectively.
% Let query token $q_i$ occupy position $(t, h, w)$ and key token $k_j$ occupy position $(t', h', w')$.
% The attention logit between the two tokens decomposes as
We show that RoPE-3D~\cite{su2024roformer, kong2024hunyuanvideo, wan2025wan} introduces structured periodicity by 
partitioning embedding channels into three disjoint subsets $M_t$, 
$M_h$, $M_w$ encoding temporal, height, and width positions.
For query $q_i$ at position $(t, h, w)$ and key $k_j$ at $(t', h', w')$, the attention logit decomposes as
\begin{align}
    a_{(t,h,w),(t',h',w')}
    =\;& \sum_{m \in M_t}
         \bigl\|q_i^{(m)}\bigr\|\,\bigl\|k_j^{(m)}\bigr\|
         \cos\Bigl(\phi^{(m)} + (t-t')\,\theta_m\Bigr)
    \notag\\
    +\;& \sum_{m \in M_h}
         \bigl\|q_i^{(m)}\bigr\|\,\bigl\|k_j^{(m)}\bigr\|
         \cos\Bigl(\phi^{(m)} + (h-h')\,\theta_m\Bigr)
    \notag\\
    +\;& \sum_{m \in M_w}
         \bigl\|q_i^{(m)}\bigr\|\,\bigl\|k_j^{(m)}\bigr\|
         \cos\Bigl(\phi^{(m)} + (w-w')\,\theta_m\Bigr),
    \label{eq:rope3d}
\end{align}
where $q_i^{(m)}, k_j^{(m)} \in \mathbb{R}^2$ are the content query and key slices for channel $m$; $\phi^{(m)}$ is the angle between $q_i^{(m)}$ and $k_j^{(m)}$; and $\theta_m = c^{-2m/d}$ is the angular frequency of channel $m$, with $d$ the
total embedding dimension and $c$ the RoPE base.
The periodicity becomes most pronounced when a single channel dominates its group~\cite{yang2026attentionpatternsexistunifying}.
Suppose a single channel $m^*_\lambda \in M_\lambda$ dominates its group, i.e.,
\begin{equation}
\bigl\|q_i^{(m^*_\lambda)}\bigr\|\,\bigl\|k_j^{(m^*_\lambda)}\bigr\|
    \;\gg\;
    \bigl\|q_i^{(m)}\bigr\|\,\bigl\|k_j^{(m)}\bigr\|
    \quad \forall\, m \in M_\lambda,\; m \neq m^*_\lambda.
    \label{eq:dominance}
\end{equation}
We empirically verify this dominance assumption on Wan~2.1, where $87.55\%$ of attention heads exhibit a dominant channel (see Appendix~\ref{sec:massive_m} for details).
Then the contribution of group $\lambda\in \{t, h, w\}$ to~\cref{eq:rope3d} reduces to a single
cosine term. Taking $\lambda = t$ as a representative example,
\begin{equation}
    a_{(t,h,w),(t',h',w')}
    \;\approx\;
    \bigl\|q_i^{(m^*_t)}\bigr\|\,\bigl\|k_j^{(m^*_t)}\bigr\|
    \cos\Bigl(\phi^{(m^*_t)} + (t-t')\,\theta_{m^*_t}\Bigr),
    \label{eq:dominant_t}
\end{equation}
and analogously for $\lambda \in \{h, w\}$ with $(t - t')$ replaced by $(h - h')$ or
$(w - w')$, respectively.
The logit is thus a cosine of the relative offset along axis $\lambda$, yielding an
axis-wise attention period of
\begin{equation}
    \mathcal{T}_\lambda
    \;=\;
    \frac{2\pi}{\theta_{m^*_\lambda}}
    \;=\;
    2\pi\,c^{\,2m^*_\lambda/d},
    \qquad \lambda \in \{t, h, w\}.
    \label{eq:period}
\end{equation}
In the heatmap visualization, this cosine periodicity means that high attention weights recur at fixed intervals $\mathcal{T}_\lambda$---producing the \textbf{stride} of the observed stripes.
Moreover, the content-dependent phase $\phi^{(m^*_\lambda)}$ can shift the cosine peak away from the main diagonal, producing the \textbf{offset}.

\noindent
\textbf{Joint effect.}
When $\mathbf{A}$ is visualized as a 2D heatmap under the flattened 1D token ordering,
the two properties identified above---video redundancy and axis-wise cosine periodicity---jointly produce the observed periodic diagonal stripes:
\textbf{(1) Inter-frame stripes} are driven by temporal locality in~\cref{eq:locality}:
the stripe \emph{width} is governed by $\varepsilon_t$, and the stripe \emph{period} is governed
by $\mathcal{T}_t$.
\textbf{(2) Intra-frame stripes} are driven by spatial locality within each frame block:
the stripe \emph{width} is governed by $\varepsilon_w$ (row-wise locality), and the stripe \emph{period} is governed by $\mathcal{T}_h$ and $\mathcal{T}_w$ (\cref{eq:period}),
with column-wise locality ($\varepsilon_h$) inducing repeated off-diagonals at intervals of $W$ under row-major flattening.

\begin{figure}[t!]
  \centering
  \includegraphics[width=1\linewidth]{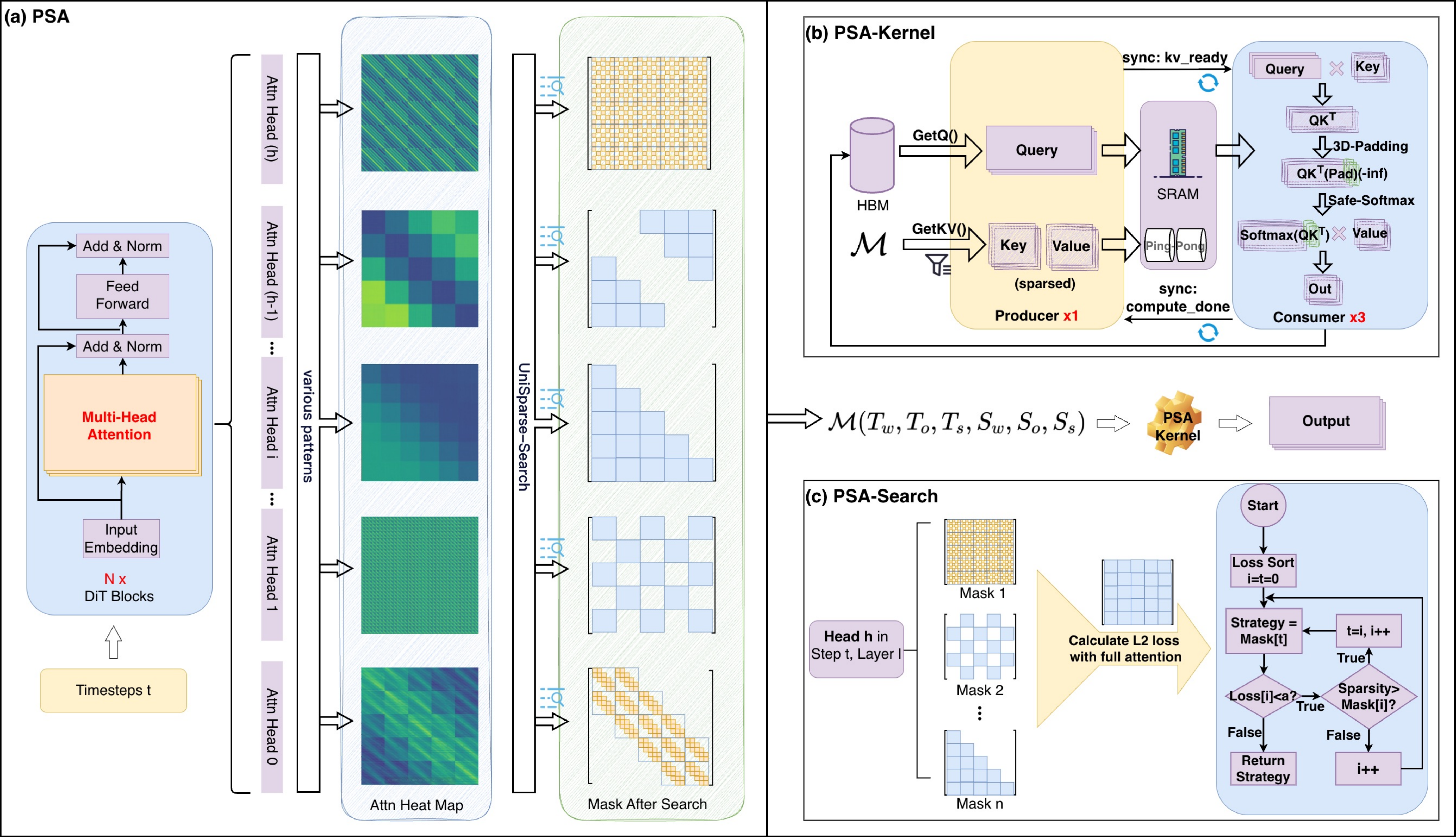}
  
  \caption{An overview of the \texttt{PSA} framework. 
    \textbf{(a) Unified Representation:} The parametric model $\mathcal{M}$ captures diverse per-head sparse attention patterns, covering both intra-frame and inter-frame structures with varying widths, offsets, and strides. 
    \textbf{(b) Efficient Kernel:} A dedicated CUDA kernel enables high-performance, hardware-aware computation across all supported sparse patterns. 
    \textbf{(c) Automatic Search:} \texttt{PSA-Search} automatically identifies the optimal sparse configuration for each head by maximizing sparsity subject to a predefined output degradation threshold.}

  \label{fig:method-main}
\end{figure}

\subsection{Efficient and Flexible Kernel}

\textbf{Unified Representation.} To efficiently encode the sparse diagonal-stripe patterns identified above and enable a single, hardware-efficient kernel, we propose a parametric formulation
$\mathcal{M}(T_w, T_o, T_s, S_w, S_o, S_s)$, where six integer parameters govern
inter-frame and intra-frame diagonal bands via their widths, offsets, and strides,
respectively, as illustrated in \cref{fig:param_vis}.
SVG and STA are special cases: SVG corresponds to
$\mathcal{M}(T_w, 0, \infty, S_w, 0, \infty)$, and STA's sliding window to
$\mathcal{M}(T_w, 0, \infty, S_w, 0, S_s)$, where $T_s = \infty$ or $S_s = \infty$
reduces the corresponding axis to a single diagonal.
These restricted configurations fail to capture multi-diagonal, hybrid, and uniform
patterns, which account for over 60\% of attention heads in HunyuanVideo and Wan~2.1.
Our formulation expresses all four observed patterns (see \cref{fig:method-main}~(a)),
achieving an average cosine similarity of 0.9831 with full attention across all heads.
To generate attention masks, we devise the generation algorithm (\cref{alg:unimask}). Given an input sequence of length $L = T \times H \times W$, the mask generation
algorithm permits an attention connection between query $q_i$ and key
$k_j$ only when both the inter-frame and intra-frame conditions of $\mathcal{M}$ are
satisfied.

\begin{algorithm}[t!]
\caption{\texttt{PSA} Mask Definition}
\label{alg:unimask}
\begin{algorithmic}[1]
\Statex \textbf{Input:} $Q$ index $i$, $K$ index $j$, Parameters $\mathcal{M}(T_w, T_o, T_s, S_w, S_o, S_s)$
\Statex \textbf{Output:} Boolean indicating if $q_i$ attends to $k_j$

\Function{IsAttending}{$i, j, \mathcal{M}$}
    \State $(t_q, s_q) \gets (\lfloor i / (H \cdot W) \rfloor, i \%(H \cdot W))$
    \State $(t_k, s_k) \gets (\lfloor j / (H \cdot W) \rfloor, j \%(H \cdot W))$

    \State $\Delta_t \gets t_k - t_q$ \Comment{inter-frame distance}
    \State $\mathrm{inter\_frame\_cond} \gets ((|\Delta_t - T_o|) \mod T_s) \le T_w-1$
    
    \If{not $\mathrm{inter\_frame\_cond}$}
        \State \Return \textbf{false}
    \EndIf

    \State $\Delta_s \gets s_k - s_q$ \Comment{intra-frame distance}
    \State $\mathrm{intra\_frame\_cond} \gets ((|\Delta_s - S_o|) \mod S_s) \le S_w-1$
    
    \State \Return $\mathrm{intra\_frame\_cond}$
\EndFunction
\end{algorithmic}
\end{algorithm}

\textbf{Hardware-Aware Design.} Our kernel design targets maximum computational efficiency
on modern GPUs by ensuring full utilization of every
scheduled thread block, eliminating idle compute units~\cite{sta}.
To this end, we introduce a fundamental computational abstraction called \texttt{PSABlock}, which represents the minimum data granularity necessary to fully saturate the resources of a single streaming multiprocessor (SM).
Notably, \texttt{PSABlock} also aligns with the block-wise characteristics observed in the attention heatmaps (\cref{fig:heatmap}), enabling the mask to operate naturally at this granularity.
Since \texttt{PSABlock} is the minimum computable unit, \texttt{PSA} applies masking at \texttt{PSABlock} granularity rather than individual tokens. For video tokens organized in 3D structure, a \texttt{PSABlock} comprises constituent blocks along temporal, height, and width dimensions: $\text{\texttt{PSABlock}} = \text{\texttt{PSABlock}}_{\text{t}} \times \text{\texttt{PSABlock}}_{\text{h}} \times \text{\texttt{PSABlock}}_{\text{w}}$. 
% Temporal mask parameters $(T_w, T_o, T_s)$ operate at the granularity of $\text{\texttt{PSABlock}}_{\text{t}}$, while spatial parameters $(S_w, S_o, S_s)$ apply to the 2D plane formed by $\text{\texttt{PSABlock}}_{\text{h}} \times \text{\texttt{PSABlock}}_{\text{w}}$. 
% This block-level masking strategy enables complex sparse patterns while ensuring computations operate on dense, hardware-friendly data chunks that preserve high throughput.

\noindent
\textbf{High-Performance Kernel.} We develop a high-performance CUDA kernel, 
\texttt{PSA-Kernel}, specifically optimized for the \texttt{PSA} 
framework on the NVIDIA Hopper GPU architecture. Built upon the 
\texttt{ThunderKittens}~\cite{spector2024thunderkittenssimplefastadorable} 
library, \texttt{PSA-Kernel} draws inspiration from its highly efficient 
FlashAttention-3 pipeline design, as illustrated in \cref{fig:method-main}~(b).

The key challenge lies in maintaining high computational efficiency across diverse sparse patterns. We address this through two synergistic mechanisms. First, we reorganize the $Q, K, V$ data layout from the standard $(T,H,W)$ format to a \texttt{PSABlock}-centric arrangement: $(T', H', W',$
${PSABlock}_t$, ${PSABlock}_h$, ${PSABlock}_w)$, where $(T', H', W')$ denotes the grid of \texttt{PSABlock}s. This transformation ensures contiguous memory placement within each \texttt{PSABlock}, enabling coalesced memory accesses---a prerequisite for the efficient kernel design that follows.
Second, building on this layout, our kernel achieves FA3-level efficiency while supporting flexible masks as shown in \cref{fig:method-main}(b).
A dedicated \emph{producer} warpgroup guided by the mask logic (\cref{alg:unimask}) identifies required $K, V$ \texttt{PSABlock}s and asynchronously prefetches them into on-chip SRAM via TMA-driven ping-pong buffering. Concurrently, \emph{consumer} warpgroups perform dense attention computations at \texttt{PSABlock} granularity using resident $Q$ and prefetched $K, V$ data. This decoupling of data movement and computation enables hardware-optimized dense operations, maximizing GPU utilization even under sparse attention patterns.

\noindent
\textbf{3D-Padding for Resolution Flexibility.} The block-aligned computation described above requires that the token dimensions $(T, H, W)$ be divisible by the corresponding \texttt{PSABlock} tile sizes $({PSABlock}_t,\allowbreak {PSABlock}_h,\allowbreak {PSABlock}_w)$. To support arbitrary video resolutions, we employ \textbf{3D-padding}: each dimension is minimally padded to the smallest multiple of its corresponding \texttt{PSABlock} size, i.e., $T' = \lceil T / {PSABlock}_t \rceil \cdot {PSABlock}_t$, ensuring alignment without restricting input geometry.
% The handling of these padded tokens is managed directly within the attention computation. After the query-key dot-product matrix ($Q \cdot K^T$) is computed, but before the softmax operation, the kernel identifies the scores corresponding to any padded token positions. These scores are then masked by overwriting them with a large negative value (i.e., negative infinity) to ensure the right safe-softmax results. This approach allows us to support arbitrary video resolutions while introducing only a minimal computational overhead. For example, at 720p resolution, padding the height dimension (H) of tokens from 45 to 48 for the final 80\% of generation steps incurs an overhead of only 5\% ($\text{Overhead} = p_{\text{steps}} \times r_{\text{padding}}$, where $p_{\text{steps}}$ is the proportion of steps where padding is applied, and $r_{\text{padding}}$ is the relative padding overhead).
% 润色：
% The handling of padded tokens is integrated directly into the attention computation. Specifically, after computing the Q-K dot-product matrix ($Q \cdot K^T$) but prior to applying the softmax operation, the kernel identifies scores corresponding to padded token positions. These scores are then masked by setting them to negative infinity, ensuring correct softmax normalization. 
This design enables support for arbitrary video resolutions with minimal computational overhead. For instance, at 720p resolution, padding the height dimension (H) from 45 to 48 tokens during the last 80\% of generation steps incurs only 5\% overhead ($\text{Overhead} = p_{\text{steps}} \times r_{\text{padding}}$, where $p_{\text{steps}}$ denotes the proportion of steps requiring padding and $r_{\text{padding}}$ represents the relative padding ratio).

\subsection{Automatic and Efficient Sparsity Discovery}
\label{subsec:psa_search}

\begin{wrapfigure}[11]{r}{0.4\linewidth}
  \vspace{-1.8em}
  \centering
  \includegraphics[width=\linewidth]{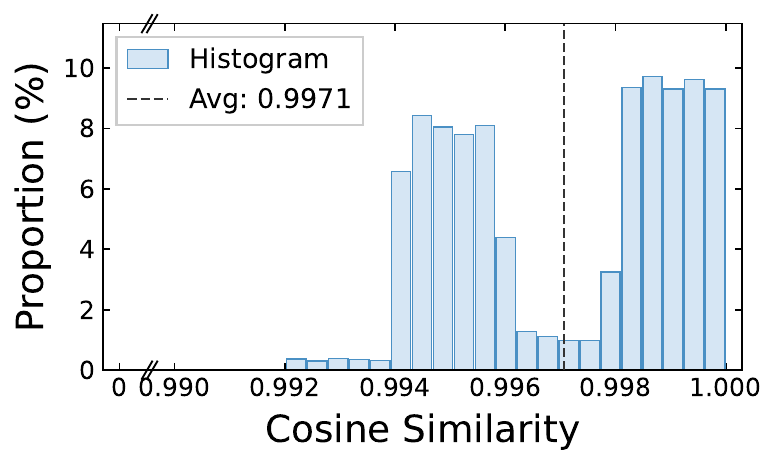}
  \captionsetup{font=scriptsize, width=\linewidth, skip=2pt}
  \caption{Evaluation of the similarity of attention heatmaps across all 11 evaluation dimensions of VBench, where prompts 1–99 are compared against prompt 0 at the same head positions within a search space of $100\ \text{(timesteps)} \times 40\ \text{(layers)} \times 40\ \text{(heads)}$.}
  \label{fig:cosine_prompt}
\end{wrapfigure}

A central challenge in applying the \texttt{PSA} representation is assigning an optimal mask configuration $\mathcal{M}$ to each attention head. We address this with \texttt{PSA-Search}, an automatic search algorithm grounded in a key empirical observation: attention patterns in DiT models vary significantly across timesteps, layers, and heads, yet remain remarkably stable with respect to input prompts~\cite{sparse_vdit} (see \cref{fig:cosine_prompt} and Appendix~\ref{sec:heat_map_diff_prompt}). This prompt-invariance enables an offline search strategy---optimal sparse configurations can be identified once using a representative dataset and subsequently applied to all future inferences, introducing no runtime overhead.

The \texttt{PSA-Search} algorithm operates on a per-head basis, as illustrated in \cref{fig:method-main}~(c). For each attention head, the algorithm iterates through a predefined search space of candidate sparse configurations. Given a user-defined \( L_2 \) loss threshold \( \alpha \), the search first identifies all candidate masks satisfying this quality constraint (i.e., \( L_2 \leq \alpha \)), where the \( L_2 \) loss is computed between the outputs of the sparse attention and the full attention. Among these qualified candidates, \texttt{PSA-Search} selects the mask achieving maximum sparsity.
Furthermore, \texttt{PSA-Search} supports a dual-interface: users can either specify an error tolerance \( \alpha \) to maximize sparsity, or specify a target sparsity level and obtain the corresponding mask with minimal error.
% To enhance search efficiency, we incorporate two optimizations: (1) an $EarlyStop$ mechanism that terminates the search when further exploration is unlikely to yield better candidates, and (2) distributed parallelization across multiple GPUs through Ulysses~\cite{jacobs2023deepspeedulyssesoptimizationsenabling}. 
% With these optimizations, searching over 100 candidate masks completes in merely 2.5 hours on 8 H100 GPUs, enabling globally optimized model configurations without the prohibitive costs associated with training-based search methods.
To enhance search efficiency, we leverage distributed parallelization across multiple GPUs via Ulysses~\cite{jacobs2023deepspeedulyssesoptimizationsenabling}. Searching over 100 candidate masks completes in 2.5 hours on 8 H100 GPUs, assigning each head its maximum-sparsity mask under the error constraint without the prohibitive costs of training-based methods.

%% file: sec/4_experiment.tex
\section{Experiments and Results}
\label{sec:experiment}

%-------------------------------------------------------------------------

\begin{table}[t!]
\centering
\caption{Comparative evaluation of \texttt{PSA} against SVG, SVG2 and STA baselines across different models. Spar denotes the average sparsity across all attention heads. ``-'' indicates metrics that are self-referential for the baseline.}
\label{tab:main_results}
\begin{adjustbox}{max width=\linewidth}
\begin{tabular}{c|ccccccccc}
\hline
\textbf{Models} & \textbf{Methods} & \textbf{Spar (\%)} & \textbf{E2E Speed (s)$\downarrow$} & 
 \textbf{OC$\uparrow$} & \textbf{TS$\uparrow$} & \textbf{MSE$\downarrow$} & 
\textbf{PSNR$\uparrow$} & \textbf{SSIM$\uparrow$} & \textbf{LPIPS$\downarrow$} \\
\hline
\multirow{5}{*}{\textbf{\shortstack{HunyuanVideo\\(720×1280)}}} & FA2 & 0&1793.06&26.95&24.52&-&-&-&- \\
& FA3 & 0&1172.62 (1x)&26.95&24.52&-&-&-&-  \\
 % & STA & / & / &/ &/ &/ &/ &/ &/ &/ \\
 & SVG & 56.92 & 906.12 (1.29x) & 22.32 & 21.11 & 0.0039 & 24.2716 & 0.8686 & 0.3400 \\
& SVG2 & 61.92 & 866.10 (1.35x) & 25.96 & 23.28 & 0.0016 & 29.6303 & 0.8910 & 0.1625 \\

 & \textbf{Ours} & \textbf{59.74} & \textbf{777.28 (1.51x)} &  \textbf{26.00} & \textbf{23.63} & \textbf{0.0012} & \textbf{30.3463} & \textbf{0.9062} & \textbf{0.1523}\\
\hline
\multirow{6}{*}{\textbf{\shortstack{HunyuanVideo\\(768×1280)}}}  & FA2 & 0&1684.60&26.97&24.61&-&-&-&- \\
& FA3 & 0&1106.36 (1x)&26.97&24.61&-&-&-&-  \\
& STA & 58.37 & 746.73 (1.48x) & 26.19 & 23.39 & 0.0596 & 12.5239 & 0.6036 & 0.3117 \\
& SVG & 56.98 & 797.57 (1.39x) & 23.21 & 22.43 & 0.0053 & 22.9984 & 0.8654 & 0.1907 \\
& SVG2 & 63.23 & 773.47 (1.43x) & 25.31 & 22.74 & 0.0027 & 25.7979 & 0.7855 & 0.2184 \\
& \textbf{Ours} & \textbf{60.33} & \textbf{703.91 (1.57x)} &  \textbf{26.33} & \textbf{23.64} & \textbf{0.0019} & \textbf{27.1983} & \textbf{0.8287} & \textbf{0.1568} \\
\hline
\multirow{5}{*}{\textbf{\shortstack{Wan~2.1 14B\\(720×1280)}}} & FA2 & 0&1873.32&25.91&23.19&-&-&-&- \\
& FA3 & 0&1214.69 (1x) &25.91&23.19&-&-&-&-  \\
 % & STA & / & / &/ &/ &/ &/ &/ &/ &/ \\
& SVG & 58.51 & 944.81 (1.29x) & 25.29 & 20.32 & 0.0157 & 18.5862 & 0.7056 & 0.2881 \\
& SVG2 & 63.40 & 926.78 (1.31x)   & 24.69 & 23.33 & 0.0051 & 23.4689 & 0.8391 & 0.1703 \\
& \textbf{Ours} & \textbf{57.57}  & \textbf{886.64 (1.37x)} &  \textbf{25.29 }& \textbf{23.59} & \textbf{0.0048} & \textbf{23.4813} & \textbf{0.8372} & \textbf{0.1601} \\
\hline
\end{tabular}
\end{adjustbox}
\end{table}

\subsection{Setup}

\textbf{Models.} We evaluate two state-of-the-art text-to-video models: \textbf{Wan~2.1} (14B parameters) with 81-frame sequences at 720×1280, and \textbf{HunyuanVideo} in two configurations—129 frames, 720×1280 and 117 frames, 768×1280.

\noindent
\textbf{Metrics.} For quantitative evaluation of frame-level fidelity and perceptual quality, we report Mean Squared Error (MSE), Peak Signal-to-Noise Ratio (PSNR), Structural Similarity Index Measure (SSIM), and Learned Perceptual Image Patch Similarity (LPIPS)~\cite{zhang2018perceptualLPIPS}. 
In addition to these per-frame metrics, we conduct a more holistic assessment using the VBench benchmark~\cite{zhang2024evaluationagentVBench}. 
Specifically, we report scores for overall consistency (\texttt{OC}) and temporal style adherence (\texttt{TS}). 

\noindent
\textbf{Baselines.} We compare against three sparse attention methods:  
\textbf{SVG} and \textbf{SVG2}, evaluated across all models to assess generalizability; 
and \textbf{STA}, evaluated only on HunyuanVideo at 768×1280 resolution due to its tile-based architectural constraints~\cite{sta}.  
% Among them, \textbf{SVG} represents a fixed-pattern sparse attention method, \textbf{STA} is a locality-based sparse approach, and \textbf{SVG2} is a top-$k$ sparse attention method that achieves state-of-the-art performance.
These methods represent the two paradigms discussed in \cref{sec:intro}: \textbf{SVG} and \textbf{STA} employ predefined masks with limited flexibility, while \textbf{SVG2} determines masks dynamically at runtime, sacrificing hardware efficiency for flexibility.

% \noindent
% Furthermore, to preserve the foundational structure of the generated content, we exempt the initial 20\% of diffusion steps from any sparsification. This strategy is motivated by the widely recognized importance~\cite{svg, svg2, xia2025trainingfreeadaptivesparseattentionadaspa, adnan2025foresightadaptivelayerreuse,chen2025rainfusionadaptivevideogeneration,zhang2025trainingfreeefficientvideogeneration} of these early steps in establishing the coarse layout and overall quality of the final output.

\noindent
\textbf{Parameters.} All methods target approximately 60\% sparsity. For \texttt{PSA}, we construct a discrete candidate space for
$\mathcal{M}$ based on attention heatmap analysis, uniformly sampling about 100 configurations per head. The threshold
$\alpha$ is set to 0.03, 0.018, and 0.007 for HunyuanVideo (720$\times$1280), HunyuanVideo (768$\times$1280), and
Wan~2.1, respectively. Search uses 1–3 prompts (inter-prompt similarity is high) and evaluation uses all VBench       
prompts. The initial 20\% of diffusion timesteps are exempted from sparsification, as early steps are critical for
coarse layout and quality~\cite{svg, svg2, xia2025trainingfreeadaptivesparseattentionadaspa,
adnan2025foresightadaptivelayerreuse, chen2025rainfusionadaptivevideogeneration,
zhang2025trainingfreeefficientvideogeneration}.

\subsection{PSA Performance and Video Quality}

\begin{table}[t!]
\centering
\caption{Kernel performance of \texttt{PSA} across different models with the same mask pattern. Spar denotes the kernel sparsity of each individual attention head.}
\label{tab:kernel_performance}
\begin{adjustbox}{max width=\linewidth}
\footnotesize
\begin{tabular}{c|ccccc}
\hline
\textbf{Models} & \textbf{Methods} & \textbf{Spar (\%)} & \textbf{Kernel Speed (ms)$\downarrow$} & \textbf{TFLOPs} & \textbf{MFU (\%)$\uparrow$} \\
\hline
\multirow{4}{*}{{\shortstack{HunyuanVideo\\(720×1280)}}} 
% & FA2 & 0&468.00 & 173.43 & 37.47  \\
& FA3 & 0&251.30 (1.00x) & 173.43 & 69.78  \\
 % & STA & / & /  \\
 & SVG & 65.41 & 173.40 (1.45x) & 59.99 & 34.98  \\
 & SVG2 & 65.67 &  156.76 (1.60x) &59.54 & 38.40 \\
 & \textbf{Ours} & \textbf{66.00} & \textbf{121.58 (2.07x)} & 58.96 & \textbf{49.04} \\
\hline
\multirow{5}{*}{{\shortstack{HunyuanVideo\\(768×1280)}}}  
% & FA2 & 0&438.06 & 163.07 & 37.64 \\
& FA3 & 0&236.36 (1.00x) &163.07&69.76  \\
& STA & 58.33 & 111.73 (2.12x) &67.95&61.50 \\
& SVG & 58.43 & 178.59 (1.32x) &67.79 &38.38 \\
& SVG2 & 57.68 & 182.79 (1.29x) & 69.01 & 38.18\\
& \textbf{Ours} & \textbf{58.33} & \textbf{112.25 (2.11x)} &67.95&\textbf{61.21}  \\
% & SVG & 65.27 & 145.56 (1.62x) &56.64&39.34 \\
% & SVG2 & 64.19 & 139.43 (1.70x) & 58.40 & 42.35\\

% & \textbf{Ours} & \textbf{64.00} & \textbf{99.31 (2.38x)} &58.71&\textbf{59.77}  \\
\hline
\multirow{4}{*}{{\shortstack{Wan~2.1 14B\\(720×1280)}}} 
% & FA2 & 0&315.71&117.05&37.49 \\
& FA3 & 0&185.55 (1.00x)&117.05&63.78 \\
 % & STA & / & /  \\
& SVG & 75.41 & 77.61 (2.39x) &28.78&37.50 \\
& SVG2 & 74.84 & 69.86 (2.66x) &29.45&42.62 \\
& \textbf{Ours} & \textbf{75.00}  & \textbf{55.12 (3.37x)} & 29.26 & \textbf{53.68}  \\
\hline
\end{tabular}
\end{adjustbox}
\end{table}

% To quantify the computational efficiency of \texttt{PSA-Kernel}, we conduct a performance analysis at two distinct levels: kernel execution speed and end-to-end (E2E) inference latency.
% All experiments were performed on a single NVIDIA H100 GPU.
% Furthermore, to situate our performance relative to state-of-the-art dense attention implementations, we benchmark against baselines implemented with FlashAttention-2 (FA2)~\cite{dao2023flashattention2} and FlashAttention-3 (FA3)~\cite{shah2024flashattention}.

We evaluate \texttt{PSA-Kernel} at both kernel and end-to-end (E2E) levels on a single NVIDIA H100 GPU, benchmarking against FlashAttention-2 (FA2)~\cite{dao2023flashattention2} and FlashAttention-3 (FA3)~\cite{shah2024flashattention}.

\noindent
\textbf{E2E inference performance.} 
% These results collectively validate that our proposed strategy achieves high execution efficiency, leading to substantial acceleration in practical end-to-end video generation tasks.
As presented in \cref{tab:main_results}, \texttt{PSA} consistently outperforms existing sparse attention baselines across different model architectures. 
% Compared to SVG and SVG2, it reduces E2E latency by \textbf{13\%} and \textbf{10\%}, respectively, and delivers a \textbf{6\%} improvement over the tile-based STA method on HunyuanVideo ($768 \times 1280$). 
 Compared to SVG, SVG2 and STA, it achieves relative speedups of \textbf{1.13$\times$}, \textbf{1.10$\times$}, and \textbf{1.06$\times$} on HunyuanVideo ($768 \times 1280$), respectively.
% We attribute SVG2's inferior performance to three limitations: inability to leverage TMA due to its reliance on FlashInfer~\cite{ye2025flashinferefficientcustomizableattention}, irregular memory access patterns from token-wise clustering, and additional clustering overhead. 

% The results summarized in \cref{tab:kernel_performance} reveals a substantial performance advantage for our implementation.
% This considerable speedup highlights the efficiency of our kernel's design in handling the adaptive sparse patterns generated by our method.
% Furthermore, while the performance gap varies by configuration, our kernel also consistently outperforms the STA implementation, as shown in the table. 
% This kernel-level efficiency is the foundational driver of the end-to-end latency reductions, confirming that our method's performance gains are rooted in a highly optimized computational core.

\noindent
\textbf{Kernel performance.}
% Notably, compared to the SVG kernel, we demonstrate a significant performance improvement of approximately \textbf{98\%} in Wan~2.1.
% Moreover, our kernel delivers performance competitive with STA.
% This kernel-level efficiency serves as the foundational driver of the end-to-end latency reductions, confirming that our method's performance gains stem from a highly optimized computational core.
To isolate the computational efficiency of our core attention mechanism, we conduct detailed micro-benchmarks of \texttt{PSA-Kernel} against the kernels from SVG, SVG2 and STA under the same mask pattern, as shown in \cref{tab:kernel_performance}. Implementation differences across methods may cause slight sparsity variations even under the same mask pattern.
% Notably, compared to the SVG and SVG2 kernel, our kernel achieves a significant speedup of approximately \textbf{47\%} and \textbf{40\%} on HunyuanVideo. 
Notably, compared to the SVG and SVG2 kernels, our kernel achieves significant speedups of approximately \textbf{1.59$\times$} and \textbf{1.63$\times$} on HunyuanVideo (768$\times$1280), respectively.
Moreover, it achieves kernel performance competitive with STA while delivering superior E2E speed---the E2E advantage stems from our FA3-based implementation for unsparsified timesteps (the initial 20\% of diffusion steps), whereas STA relies on a slower custom full-attention kernel during these steps.
This kernel-level efficiency serves as the foundational driver of the end-to-end latency reductions, confirming that our method's performance gains stem from a highly optimized computational core.

% \subsection{Video Quality}

\noindent
\textbf{Video Quality.} Beyond computational performance, a critical question is whether \texttt{PSA} preserves video generation quality. As shown in \cref{tab:main_results}, \texttt{PSA} achieves the best VBench scores (OC and TS) among all sparse methods across all configurations, while also delivering the best frame-level fidelity metrics (MSE, PSNR, SSIM, LPIPS). Notably, on HunyuanVideo (720$\times$1280), \texttt{PSA} achieves a PSNR of 30.35 at 59.74\% sparsity, confirming that the parameterized mask $\mathcal{M}$ faithfully captures each head's attention structure.

% It achieves highly competitive scores for MSE, PSNR, SSIM, and LPIPS, and this demonstrates that our parameterized mask $\mathcal{M}(T_w,T_o,T_s, S_w,S_o,S_s)$ can better adapt to the sparse patterns of each head.

\subsection{Sensitivity Tests}

\textbf{Quality sensitivity tests}. As shown in \cref{tab:metrics_sensitivity}, we evaluate the generation quality metrics under varying levels of sparsity. The results clearly indicate that as sparsity increases from 30\% to 60\%, all evaluation metrics exhibit a smooth and controlled degradation. 
% Specifically, the error-based metrics MSE and LPIPS gradually increase, while the quality-based metrics PSNR and SSIM decrease accordingly. This aligns with the expectation that sparsity trades some detail for efficiency.
Notably, even at a high sparsity of 60\%, the model maintains an acceptable generation quality (e.g., PSNR remains above 30). This demonstrates that the negative impact on generation quality is within a manageable range, even while unlocking significant computational savings.

% \begin{table}[h!]
% \centering
% \caption{Quality Sensitivity Tests. (in HunyuanVideo $720\times1280$)}
% \label{tab:metrics_sensitivity}
% \begin{adjustbox}{max width=\linewidth}
% \begin{tabular}{ccccc}
% \toprule
% \textbf{Sparsity} & {\textbf{MSE}$\downarrow$} & {\textbf{PSNR}$\uparrow$} & {\textbf{SSIM}$\uparrow$} & {\textbf{LPIPS}$\downarrow$} \\
% \midrule
% 30\% & 0.0005 & 34.9341 & 0.9519 & 0.0834 \\
% 40\% & 0.0007 & 33.7610 & 0.9431 & 0.0875 \\
% 50\% & 0.0010 & 31.6886 & 0.9171 & 0.1250 \\
% 60\% & 0.0012 & 30.3463 & 0.9062 & 0.1523 \\
% \bottomrule
% \end{tabular}
% \end{adjustbox}
% \end{table}

\noindent
\textbf{Kernel sensitivity tests}. \Cref{tab:kernel_performance_sensitivity} presents kernel performance across varying sparsity levels. Our kernel consistently outperforms SVG across all sparsity settings. While STA exhibits a marginal latency advantage at the kernel level, our kernel supports arbitrary video resolutions via 3D-padding with negligible overhead, whereas STA is constrained to a single fixed resolution. This generality, combined with competitive kernel throughput, gives our method practical deployment value. 

% \begin{table}[h!]
% \centering
% \caption{Kernel Performance Sensitivity Tests (ms, in HunyuanVideo $768\times1280$)}
% \label{tab:kernel_performance_sensitivity}
% \begin{adjustbox}{max width=\linewidth}
% \small
% \begin{tabular}{ccccc}
% \toprule
% \textbf{Sparsity} & \textbf{FA3 Baseline} & \textbf{STA} & \textbf{SVG} & \textbf{Ours} \\
% \midrule
% 50\% & 236.366 & 138.201 & 205.714 & 139.993 \\
% 60\% & 236.366 & 111.508 & 165.028 & 113.494 \\
% 70\% & 236.366 & 79.702  & 125.161 & 80.438  \\
% 80\% & 236.366 & 53.624  & 85.027  & 54.093  \\
% 90\% & 236.366 & 25.882  & 45.643  & 26.292  \\
% \bottomrule
% \end{tabular}
% \end{adjustbox}
% \end{table}

\begin{table}[t!]
\centering
\caption{Sensitivity tests on quality and kernel performance.}
\label{tab:sensitivity}
\begin{subtable}[t]{0.48\linewidth}
\centering
\caption{Quality (HunyuanVideo $720\times1280$)}
\label{tab:metrics_sensitivity}
\begin{adjustbox}{min width=\linewidth, max width=\linewidth}
{\renewcommand{\arraystretch}{0.98}
\begin{tabular}{ccccc}
\toprule
\textbf{Sparsity} & {\textbf{MSE}$\downarrow$} & {\textbf{PSNR}$\uparrow$} & {\textbf{SSIM}$\uparrow$} & {\textbf{LPIPS}$\downarrow$} \\
\midrule
30\% & 0.0005 & 34.9341 & 0.9519 & 0.0834 \\
40\% & 0.0007 & 33.7610 & 0.9431 & 0.0875 \\
50\% & 0.0010 & 31.6886 & 0.9171 & 0.1250 \\
60\% & 0.0012 & 30.3463 & 0.9062 & 0.1523 \\
\bottomrule
\end{tabular}}
\end{adjustbox}
\end{subtable}
\hfill
\begin{subtable}[t]{0.48\linewidth}
\centering
\caption{Performance (ms, HunyuanVideo $768\times1280$)}
\label{tab:kernel_performance_sensitivity}
\begin{adjustbox}{min width=\linewidth, max width=\linewidth}
\begin{tabular}{ccccc}
\toprule
\textbf{Sparsity} & \textbf{FA3} & \textbf{STA} & \textbf{SVG} & \textbf{Ours} \\
\midrule
% 50\% & 236.366 & 138.201 & 205.714 & 139.993 \\
60\% & 236.366 & 111.508 & 165.028 & 113.494 \\
70\% & 236.366 & 79.702  & 125.161 & 80.438  \\
80\% & 236.366 & 53.624  & 85.027  & 54.093  \\
90\% & 236.366 & 25.882  & 45.643  & 26.292  \\
\bottomrule
\end{tabular}
\end{adjustbox}
\end{subtable}
\end{table}

\subsection{Bracketing Comparison}
\label{subsec:bracketing}

For all baselines, sparsity is an \emph{outcome} rather than an \emph{input}: STA's is fixed by its predefined window masks, SVG's is quantized to the nearest feasible window combination, and SVG2's emerges from data-dependent runtime clustering, so an exactly matched sparsity budget cannot be enforced on the baseline side. We therefore evaluate two operating points on HunyuanVideo ($768\times1280$) whose sparsity \emph{strictly brackets} all baselines: \textbf{Ours-min} below every baseline and \textbf{Ours-max} above every baseline. As shown in \cref{tab:bracketing}, PSA dominates every baseline in both speed and quality at both endpoints: Ours-min is the fastest despite computing the most (1.51$\times$ vs.\ at most 1.48$\times$), and Ours-max achieves the best PSNR despite computing the least (26.76 vs.\ 25.80), confirming that the conclusions of \cref{tab:main_results} hold regardless of any residual sparsity mismatch.

\begin{table}[t!]
\centering
\caption{Bracketing comparison on HunyuanVideo ($768\times1280$). Ours-min and Ours-max are two operating points whose sparsity strictly brackets that of all baselines.}
\label{tab:bracketing}
\footnotesize
\begin{tabular}{lccc}
\toprule
\textbf{Method} & \textbf{Spar (\%)} & \textbf{Speedup$\uparrow$} & \textbf{PSNR$\uparrow$} \\
\midrule
SVG & 56.98 & 1.39$\times$ & 23.00 \\
STA & 58.37 & 1.48$\times$ & 12.52 \\
SVG2 & 63.23 & 1.43$\times$ & 25.80 \\
\midrule
\textbf{Ours-min} & 56.47 & \textbf{1.51$\times$} & \textbf{28.73} \\
\textbf{Ours-max} & 63.84 & \textbf{1.65$\times$} & \textbf{26.76} \\
\bottomrule
\end{tabular}
\end{table}

\subsection{Generality of the Stripe Structure}
\label{subsec:generality}

To examine how broadly the structural premise underlying \texttt{PSA} holds, we measure the fraction of dominant-channel heads (\cref{eq:dominance}) on nine model settings, following the protocol of Appendix~\ref{sec:massive_m}. As shown in \cref{tab:nine_settings}, the dominant-channel fraction exceeds 78.9\% on all nine settings (median $\approx$90\%), spanning resolutions (480p/720p/768p), model sizes (14B/1.3B), architectures (MM-DiT, text--visual concatenated, and factorized spatio-temporal attention), modalities (video/image), tokenizations (2D/3D patchify), and denoising schedules (50 vs.\ 40 steps). The attention heatmaps of all these models consistently exhibit the same periodic diagonal stripes as \cref{fig:heatmap}. Notably, the RoPE-free spatial attention of Open-Sora v1.2~\cite{opensora} still exhibits locality-induced stripes with stride $W$, indicating that spatio-temporal locality (\cref{eq:locality}) holds independently of RoPE and tokenization.

\begin{table}[t!]
\centering
\caption{Fraction of dominant-channel heads across nine model settings.}
\label{tab:nine_settings}
\begin{adjustbox}{max width=\linewidth}
\footnotesize
\begin{tabular}{llc}
\toprule
\textbf{Model} & \textbf{Type} & \textbf{Dominant-channel heads (\%)} \\
\midrule
Wan~2.1 14B (50-step) & video DiT, 3D full attention & 87.55 \\
Wan~2.1 14B (40-step) & video DiT, 3D full attention & 89.30 \\
Wan~2.1 1.3B & video DiT, 3D full attention, 832$\times$480 & 90.81 \\
HunyuanVideo ($720\times1280$) & video MM-DiT & 96.61 \\
HunyuanVideo ($768\times1280$) & video MM-DiT & 96.33 \\
FLUX.1-schnell~\cite{flux2024} & image MM-DiT, 2D RoPE & 89.16 \\
CogVideoX-5b~\cite{yang2024cogvideox} & video DiT, concatenated attention & 90.39 \\
Open-Sora v1.2 & video DiT, factorized attention & 91.95 \\
Hunyuan-DiT v1.2~\cite{li2024hunyuandit} & image DiT, 2D patchify & 78.93 \\
\bottomrule
\end{tabular}
\end{adjustbox}
\end{table}

\subsection{Fidelity of Searched Masks}

% Recall that our unified representation $\mathcal{M}$ achieves an average cosine similarity of 0.9831 with full attention across all heads (\cref{sec:unified_pattern_para}). We now provide a more detailed fidelity analysis. At 60\% sparsity across all attention heads in HunyuanVideo and Wan~2.1, over 93.6\% of attention heads maintain an $L_2$ loss below $10^{-2}$ after sparsification. These results confirm that the masks identified by \texttt{PSA-Search} serve as high-fidelity approximations of the original dense attention.

Our masks achieve an average cosine similarity of 0.9831 with full attention. At 60\% sparsity, over 93.6\% of heads maintain $L_2$ loss below $10^{-2}$, confirming that \texttt{PSA-Search} produces high-fidelity sparse approximations.

\subsection{Ablation Studies}

To further validate that $\mathcal{M}(T_w, T_o, T_s, S_w, S_o, S_s)$ effectively and efficiently represents the sparse patterns of attention heads, we conduct a comprehensive ablation study.
For a fair comparison, all experiments are conducted at a fixed sparsity level, and we evaluate the quality of the generated videos.
The following five scenarios were tested:
\begin{enumerate}
\setlength{\itemsep}{1pt}
\setlength{\parsep}{0pt}
\setlength{\parskip}{0pt}
    \item \textbf{SVG-like ($\mathcal{M}(T_w, S_w)$):} This configuration simulates the approach of SVG, where only the width of the temporal and spatial attention bands can be controlled.
    
    \item \textbf{STA-like ($\mathcal{M}(T_w, S_w, S_s)$):} This mask simulates the behavior of STA. STA's method of applying sparsity along the tile's t, w, and h axes corresponds in the attention heatmap to defining the inter-frame width ($T_w$), intra-frame width ($S_w$), and the stride of intra-frame diagonals ($S_s$).
    
    \item \textbf{Inter-frame-Only ($\mathcal{M}(T_w, T_o, T_s)$)}.
    
    \item \textbf{Intra-frame-Only ($\mathcal{M}(S_w, S_o, S_s)$)}.
    
    \item \textbf{Ours ($\mathcal{M}(T_w, T_o, T_s, S_w, S_o, S_s)$)}.
\end{enumerate}

\begin{table}[t!]
\centering
\caption{Ablation study on mask at a fixed 60\% sparsity (HunyuanVideo, $720\times1280$). }
\label{tab:ablation_study}
\begin{adjustbox}{max width=\linewidth}
\begin{tabular}{lcccc}
\toprule
\textbf{Config} & {\textbf{MSE}$\downarrow$} & {\textbf{PSNR}$\uparrow$} & {\textbf{SSIM}$\uparrow$} & {\textbf{LPIPS}$\downarrow$} \\
\midrule
SVG-like & 0.0027 & 26.0230 & 0.8364 & 0.2073 \\
STA-like & 0.0026 & 26.1466 & 0.8133 & 0.2343 \\
Inter-frame & 0.0019 & 27.8065 & 0.8505 & 0.1757 \\
Intra-frame & 0.0042 & 24.0145 & 0.7838 & 0.2730 \\
\midrule
\textbf{Ours} & \bfseries 0.0012 & \bfseries 30.3463 & \bfseries 0.9062 & \bfseries 0.1523 \\
\bottomrule
\end{tabular}
\end{adjustbox}
\end{table}

% The results presented in \cref{tab:ablation_study} offer compelling insights. 
% As shown in \cref{tab:ablation_study}, the full six-parameter model outperforms all ablated versions across every metric, achieving a PSNR of 30.35---2.5 dB higher than the next best configuration (Inter-frame-only). This confirms that SVG-like and STA-like configurations, which lack offset and stride parameters, cannot adequately express the multi-diagonal and hybrid patterns. A notable asymmetry emerges between Inter-frame-only and Intra-frame-only: the former achieves a PSNR of 27.81, while the latter drops to 24.01. This gap highlights that inter-frame relationships are more critical for video generation quality, yet intra-frame patterns remain essential---only the full model captures both. It is also noteworthy that the SVG-like configuration using our method outperforms the actual SVG in \cref{tab:main_results}, as \texttt{PSA-Search} assigns more accurate per-head sparse patterns than SVG's fixed two-pattern strategy.

As shown in \cref{tab:ablation_study}, the full six-parameter model outperforms all ablated versions, achieving a PSNR of 30.35---2.5 dB higher than the next best (Inter-frame-only). This confirms that offset and stride parameters are essential for capturing multi-diagonal and hybrid patterns. Inter-frame-only (PSNR 27.81) significantly outperforms Intra-frame-only (24.01), showing that inter-frame relationships are more critical, yet only the full model captures both. Notably, our SVG-like configuration even outperforms the actual SVG in \cref{tab:main_results}, as \texttt{PSA-Search} assigns more accurate per-head patterns than SVG's fixed two-pattern strategy.

% A crucial observation arises from comparing the \textbf{Inter-frame-Only} and \textbf{Intra-frame-Only} masks. The Inter-frame-only configuration achieves a PSNR of 27.81, whereas the intra-frame-only variant drops to 24.01. This significant gap underscores the critical importance of modeling inter-frame (temporal) relationships in video generation. Furthermore, our full model substantially surpasses simplified approximations like the SVG-like and STA-like configurations. These results strongly support our central claim: all six parameters are essential for flexibly and accurately capturing the diverse attention patterns, proving the superior adaptability and expressive power of our mask design.

%% file: sec/5_conclusion.tex
\section{Conclusion}
\label{sec:conclusion}

%-------------------------------------------------------------------------

We present \texttt{PSA}, a parameterized stripe attention framework for efficient video generation. Our central insight is that video DiT attention is not arbitrary but exhibits periodic diagonal stripe structures along both temporal and spatial dimensions. This structural understanding enables a unified parametric formulation $\mathcal{M}(T_w, T_o, T_s, S_w, S_o, S_s)$ that resolves the flexibility--efficiency tension inherent in prior methods: it covers all observed attention patterns while permitting a single hardware-efficient CUDA kernel achieving up to 61\% MFU. A training-free offline search algorithm then assigns each head its maximum-sparsity mask under an error constraint, introducing no runtime overhead. Experiments on HunyuanVideo and Wan~2.1 demonstrate end-to-end speedups of 1.57$\times$ and 1.37$\times$ over FlashAttention-3 with minimal quality degradation.

%% file: sec/acknowledgements.tex
% Funding statement and disclosures, required for the camera-ready version.
\begin{ack}
This work was supported by Alibaba Group through Alibaba Research Intern Program.
Like most work on large-scale video generation, developing and evaluating our method required substantial computational resources, which carries environmental costs and is currently feasible only for research groups with access to large GPU clusters. While our method is training-free---built on top of existing pretrained video generation models without requiring additional fine-tuning---and reduces per-inference compute and energy, the overall computational demands of video generation remain significant.
\end{ack}

%% file: sec/X_suppl.tex
\clearpage

This appendix provides supplementary materials to support the main paper. We organize the content as follows:

\textbf{\Cref{sec:limitations}} discusses the limitations of the current work and outlines potential directions for future research.

\textbf{\Cref{sec:appendix_pseudocode}} presents the pseudocode implementation of \texttt{PSA-Kernel}, offering detailed algorithmic specifications for reproducibility.

\textbf{\Cref{sec:appendix_results}} introduces the \texttt{PSA-Search} process, including how to initialize mask candidates and the search results.

\textbf{\Cref{sec:locality_stripes}} provides the formal proposition and derivation showing how video spatio-temporal locality alone produces periodic diagonal stripes in the attention heatmap.

\textbf{\Cref{sec:massive_m}} provides extensive empirical evidence, through large-scale visualization experiments on Wan~2.1 and HunyuanVideo, to validate the dominance assumption regarding $m^*_\lambda$ proposed in \cref{sec:unified_pattern_para}.

\textbf{\Cref{sec:heat_map_cogvideo}} demonstrates that CogVideoX-v1.5 exhibits attention heatmap characteristics consistent with our observations, confirming that our method can be effectively applied to accelerate computation in this model as well.

\textbf{\Cref{sec:heat_map_diff_prompt}} visualizes attention heatmaps across various prompts, revealing that identical heads exhibit highly similar heatmap patterns regardless of input prompt variations.

% \textbf{\Cref{sec:Ablation_earlystop}} presents the ablation study on the $EarlyStop$ parameter, where we select appropriate values to balance sparsity and generation quality.

\textbf{\Cref{sec:Ablation_alpha}} presents the loss threshold sensitivity analysis.

\textbf{\Cref{sec:e2e-comp-diff-sparsity}} presents a comprehensive evaluation of end-to-end performance and generation quality across varying sparsity levels.

\textbf{\Cref{sec:qualitative}} provides qualitative comparisons with the FlashAttention-3 baseline, showcasing visual quality assessments and demonstrating that our method maintains comparable generation quality while achieving significant speedups.

\section{Limitations}
\label{sec:limitations}

% While \texttt{PSA} demonstrates substantial acceleration for DiTs, the current work is limited to video generation models, and it remains uncertain whether the same approach transfers effectively to other model families or domains, such as language models. Nonetheless, our preliminary analysis suggests that language models also exhibit the periodic stripe attention patterns identified in \cref{sec:unified_pattern_para}, indicating potential for broader applicability. We leave a thorough exploration of \texttt{PSA} beyond video generation as future work.

While \texttt{PSA} demonstrates substantial acceleration for DiTs, two limitations warrant discussion. First, the current work is limited to video generation models, and it remains uncertain whether the same approach transfers effectively to other model families or domains, such as language models.
% Nonetheless, our preliminary analysis suggests that language models also exhibit the periodic stripe attention patterns identified in \cref{sec:unified_pattern_para}, indicating potential for broader applicability. We leave a thorough exploration of \texttt{PSA} beyond video generation as future work. 
Second, the current \texttt{PSA-Search} requires 2.5 hours on 8 H100 GPUs to evaluate all candidate masks for every attention head, which can be costly. In future work, we plan to accelerate this process via a two-stage search strategy: a coarse-grained stage that rapidly filters out unpromising candidates, followed by a fine-grained stage that performs detailed evaluation only on the surviving mask candidates.

\section{Pseudocode for PSA-Kernel}
\label{sec:appendix_pseudocode}

\Cref{alg:psa-Kernel_conceptual} outlines the execution flow of \texttt{PSA-Kernel}, which employs a producer-consumer model to maximize hardware utilization. The workload is partitioned between two concurrently operating warpgroups within each GPU thread block. The \textbf{producer warpgroup} interprets the sparse mask configuration $\mathcal{M}$ to identify required \texttt{KV} blocks and asynchronously loads them from HBM to shared memory. The \textbf{consumer warpgroups} perform block-wise attention computation on available \texttt{KV} blocks, including safe softmax and result accumulation. Double-buffering with \texttt{kv\_ready} and \texttt{compute\_done} flags coordinates the pipeline, effectively hiding memory latency behind computation.

\begin{algorithm}[htbp]
\caption{Pseudocode for \texttt{PSA-Kernel}}
\label{alg:psa-Kernel_conceptual}
\begin{algorithmic}[1]
\Statex \textbf{Input:} Sparsity parameters $\mathcal{M}$, Tensors $Q, K, V$
\Statex \textbf{Output:} Result tensor \texttt{Out}

\Statex // \textit{Executed in parallel by dedicated warp groups within a thread block}
\Statex

\Procedure{Producer-Warpgroup}{}
    \State // \textit{Determine which KV blocks are needed for this Q block}
    \State $KV_{list} \gets \text{GetKVBlocks}(\text{$Q_{block}$}, \mathcal{M})$ \Comment{Uses logic from \cref{alg:unimask}}
    \For{each $KV_{block}$ in $KV_{list}$}
        \State \textbf{wait} until compute\_done flag is set
        \State Reset compute\_done flag
        \State Load $KV_{block}$ from HBM into SRAM
        \State Set kv\_ready flag
    \EndFor
\EndProcedure

\Statex

\Procedure{Consumer-Warpgroup}{}
    \For{$k \gets 1$ to $|KV_{list}|$} \Comment{Loop terminates after processing all blocks}
        \State \textbf{wait} until kv\_ready flag is set
        \State Reset kv\_ready flag
        \State // \textit{Perform attention on data in SRAM}
        \State $S \gets Q_{block} \times K_{SRAM}^T$ \Comment{Compute attention scores}
        \State $S_{padded} \gets \Call{3D-Padding}{S, -\infty}$ \Comment{Pad invalid positions with $-\infty$}
        \State $P \gets \Call{SafeSoftmax}{S_{padded}}$ \Comment{Apply safe softmax}
        \State $O_{partial} \gets P \times V_{SRAM}$ \Comment{Weighted sum of values}
        \State Accumulate $O_{partial}$ into local output $O_{local}$
        \State Set compute\_done flag
    \EndFor
    \State Write final accumulated $O_{local}$ to global memory \texttt{Out}
\EndProcedure

\end{algorithmic}
\end{algorithm}

\section{Process of PSA-Search}
\label{sec:appendix_results}

\subsection{Pseudocode for PSA-Search}

We present the detailed pseudocode of \texttt{PSA-Search} in \cref{alg:psa-Search_concise}. Given a list of candidate masks $\mathcal{C_{M}}$ and a user-specified loss threshold $\alpha$, the algorithm independently searches for the optimal mask configuration for each attention head. For a given head $h$, \texttt{PSA-Search} first evaluates all candidate masks by computing the $L_2$ loss between the sparse attention output and the full attention output, and sorts the results in ascending order of loss. Starting from the lowest-loss mask as the initial best candidate, the algorithm then scans through all qualified masks whose loss does not exceed $\alpha$ and greedily replaces the current best with any mask that achieves higher sparsity. This design ensures that the selected mask maximizes sparsity while remaining within the acceptable quality budget. After processing all heads, the algorithm returns a complete mask strategy map $\mathcal{S_{M}}$ that assigns the sparsified yet faithful mask to each head. Notably, since \texttt{PSA-Search} records the per-head loss for every candidate mask, adjusting $\alpha$ only requires re-executing lines~10--21 without repeating the costly $L_2$ loss evaluation, making the threshold tuning process highly efficient.

\begin{algorithm}[t!]
\caption{PSA-Search}
\label{alg:psa-Search_concise}
\begin{algorithmic}[1]
\Statex \textbf{Input:} List of candidate masks $\mathcal{C_{M}}$, DiT model $\mathcal{D}$, Loss threshold $\alpha$
\Statex \textbf{Output:} An optimized mask strategy $\mathcal{S_{M}}$

\Function{PSA-Search}{$\mathcal{D}, \mathcal{C_{M}}, \alpha$}
    \State $\mathcal{S_{M}} \gets \emptyset$ \Comment{Initialize an empty mask strategy map}
    \For{each head $h$ in model $\mathcal{D}$} \Comment{Iterate over all timesteps, layers, and heads}
        \State $HeadResults \gets []$
        \For{each mask $\mathcal{M} \in \mathcal{C_{M}}$}
            \State $Loss \gets \text{L2Loss}(\text{PSA}(h, \mathcal{M}), \text{FullAttn}(h))$
            \State Add $(Loss, \mathcal{M})$ to $HeadResults$
        \EndFor
        
        \State Sort $HeadResults$ by loss in ascending order.
        
        \State $Best\mathcal{M} \gets HeadResults[0].\text{mask}$ 
        % \State $StepCount \gets 0$
        \For{each $(L_i, \mathcal{M}_i)$ in $HeadResults$}
            % \State $StepCount ++$
            % \If{$L_i \le \alpha $ \textbf{and} $ StepCount \le EarlyStop$}
            \If{$L_i \le \alpha $}
                \If{$\text{Sparsity}(\mathcal{M}_i) > \text{Sparsity}(Best\mathcal{M})$}
                    \State $ Best\mathcal{M} \gets \mathcal{M}_i$
                \EndIf
            \EndIf
        \EndFor
        \State $\mathcal{S_{M}}[h] \gets Best\mathcal{M}$
    \EndFor
    \State \Return $\mathcal{S_{M}}$
\EndFunction
\end{algorithmic}
\end{algorithm}

\subsection{Initialization of Mask Candidates.}
Before conducting \texttt{PSA-Search}, it is necessary to select appropriate masks for initialization. We choose the corresponding masks based on visualized heatmap results (\cref{fig:heatmap}). In general, for a DiT with shape $(T, H, W)$ and \texttt{PSABlock} with shape $(\texttt{PSABlock}_t, \texttt{PSABlock}_h, \texttt{PSABlock}_w)$, we configure $\mathcal{M}(T_w, T_o, T_s, S_w, S_o, S_s)$ such that $0 < T_w < T/\text{PSABlock}_t$, $T_o \in \{0, \pm1, \pm2\}$, $T_s \in \{1, 2, 3, \infty\}$, $0 < S_w < W/\text{PSABlock}_w$, $S_o \in \{0, \pm1\}$, and $S_s \in \{1, 3, 5, 6, \infty\}$.

For example, in Wan~2.1, the DiT shape is $21\times45\times80$. After applying 3D-Padding to obtain $21\times48\times80$, we set the \texttt{PSABlock} size to $3\times8\times16$. Then, the ranges for $\mathcal{M}(T_w, T_o, T_s, S_w, S_o, S_s)$ are configured as follows: $T_w < 7$, $T_o \in \{0, \pm1, \pm2\}$, $T_s \in \{1, 2, 3, \infty\}$, $S_w < 5$, $S_o \in \{0, \pm1\}$, $S_s  \in \{1, 3, 5, 6, \infty\}$.

Similarly, for HunyuanVideo at $720\times1280$ resolution, the DiT shape is $33\times45\times80$. After 3D-Padding to $33\times48\times80$, we design the \texttt{PSABlock} as $3\times8\times16$. The ranges for $\mathcal{M}(T_w, T_o, T_s, S_w, S_o, S_s)$ are set as: $T_w < 11$, $T_o \in \{0, \pm1, \pm2\}$, $T_s \in \{1, 2, 3, \infty\}$, $S_w < 5$, $S_o \in \{0, \pm1\}$, $S_s  \in \{1, 3, 5, 6, \infty\}$.

For HunyuanVideo at $768\times1280$ resolution, the DiT shape is $30\times48\times80$ without 3D-Padding. We design the \texttt{PSABlock} as $6\times8\times8$, and the ranges for $\mathcal{M}(T_w, T_o, T_s, S_w, S_o, S_s)$ are configured as: $T_w < 5$, $T_o \in \{0, \pm1, \pm2\}$, $T_s \in \{1, 2, 3, \infty\}$, $S_w < 10$, $S_o \in \{0, \pm1\}$, $S_s  \in \{1, 3, 5, 6,  \infty\}$.

It is worth noting that setting \( T_s = \infty \) or \( S_s = \infty \) indicates that there is only one diagonal in the inter-frame or intra-frame attention, respectively. In practice, it suffices to set \( T_s \) no less than the number of frames $T$ and \( S_s \) no less than \( H \times W \).

% \subsection{Fidelity of Searched Masks}

% Our statistical analysis demonstrates that the sparse masks closely approximate full attention outputs. At 60\% sparsity across all attention heads in HunyuanVideo and Wan~2.1, the average cosine similarity with the FA3 baseline reaches \textbf{0.9831}. Moreover, over 93.6\% of attention heads maintain an L2 loss below $10^{-2}$ after sparsification. These results numerically confirm that our sparse masks serve as high-fidelity approximations of the original dense attention.

\subsection{Wan~2.1 14B}

To demonstrate the practical effectiveness of our framework, we analyzed the mask distribution on the Wan~2.1 14B model. Using a loss tolerance of $\alpha=0.007$, our search achieved a significant 57.57\% overall sparsity, as shown in \cref{tab:top_masks}.

\begin{table}[ht!]
\centering
\caption{Top 10 most frequent \texttt{PSA} configurations discovered by the search algorithm in Wan~2.1 14B using a search threshold of $\alpha=0.007$.}
\label{tab:top_masks}
\begin{tabular}{ccccccrr}
\toprule
\multicolumn{2}{c}{\textbf{Mask Configuration}} & \multicolumn{2}{c}{\textbf{Head Count}} \\
\cmidrule(lr){1-2} \cmidrule(lr){3-4}
{$(T_w, T_o, T_s)$} & {$(S_w, S_o, S_s)$} & {\# Heads} & {\% Total} \\
\midrule
\multicolumn{2}{c}{full window} & 32000 & 20.00 \\
(7, 0, $\infty$) & (3, 0, 3) & 25257 & 15.79 \\
(1, 0, $\infty$) & (5, 0, 6) & 11778 & 7.36  \\
(7, 0, $\infty$) & (1, 0, 3) & 11069 & 6.92  \\
(7, 0, $\infty$) & (3, 0, 1) & 9014  & 5.63  \\
(7, 0, $\infty$) & (3, 0, 5) & 6810  & 4.26  \\
(7, 0, $\infty$) & (1, 0, 5) & 5655  & 3.53  \\
(7, 0, $\infty$) & (5, 0, 3) & 5470  & 3.42  \\
(1, 0, 3) & (5, 0, 6) & 5088  & 3.18  \\
(3, 0, $\infty$) & (5, 0, 6) & 4622  & 2.89  \\
\midrule
\multicolumn{2}{l}{All Other Configurations} & 43237 & 27.02 \\ % Calculated sum of the rest
\midrule
\multicolumn{2}{l}{\textbf{Total}} & \bfseries 160000 & \bfseries 100.00 \\
\bottomrule
\end{tabular}
\end{table}

\subsection{HunyuanVideo 768$\times$1280}

On the HunyuanVideo model at a 768$\times$1280 resolution, we set the $L_2$ loss threshold to $\alpha=0.018$ for \texttt{PSA-Search}. This resulted in a heterogeneous mask combination that achieved an overall sparsity of 60.33\%. The distribution of the ten most prevalent mask configurations is presented in \cref{tab:hunyuan_masks}.

\begin{table}[ht!]
\centering
\caption{
    Top 10 most frequent \texttt{PSA} configurations discovered for the HunyuanVideo model (768$\times$1280 resolution) using a search threshold of $\alpha=0.018$.
}
\label{tab:hunyuan_masks}
\begin{tabular}{ccccccrr}
\toprule
\multicolumn{2}{c}{\textbf{Mask Configuration}} & \multicolumn{2}{c}{\textbf{Head Count}} \\
\cmidrule(lr){1-2} \cmidrule(lr){3-4}
{$(T_w, T_o, T_s)$} & {$(S_w, S_o, S_s)$} & {\# Heads} & {\% Total} \\
\midrule
\multicolumn{2}{c}{full window} & 14404 & 20.00 \\
(1, 0, $\infty$) & (10, 0, 6) & 9213  & 12.80 \\
(5, 0, $\infty$) & (1, 0, 1)  & 5494  & 7.63  \\
(5, 0, $\infty$) & (1, 0, 3)  & 4889  & 6.79  \\
(5, 0, $\infty$) & (3, 0, 3)  & 4651  & 6.46  \\
(5, 0, $\infty$) & (3, 0, 1)  & 3828  & 5.32  \\
(5, 0, $\infty$) & (1, 0, 5)  & 1789  & 2.48  \\
(5, 0, $\infty$) & (1, 0, 6)  & 1701  & 2.36  \\
(5, 0, $\infty$) & (5, 0, 1)  & 1567  & 2.18  \\
(5, 0, $\infty$) & (7, 0, 6)  & 1522  & 2.11  \\
\midrule
\multicolumn{2}{l}{All Other Configurations} & 22942 & 31.86 \\
\midrule
\multicolumn{2}{l}{\textbf{Total}} & \bfseries 72000 & \bfseries 100.00 \\
\bottomrule
\end{tabular}
\end{table}

\subsection{HunyuanVideo 720$\times$1280}

For the HunyuanVideo model at a 720$\times$1280 resolution, we applied PSA-Search using a loss threshold of $\alpha=0.03$. This process yielded a mask combination achieving 59.74\% overall model sparsity. \cref{tab:hunyuan720p_masks} summarizes the ten most frequently chosen mask configurations discovered during this search.

\begin{table}[ht!]
\centering
\caption{
    Top 10 most frequent \texttt{PSA} configurations for the HunyuanVideo model (720$\times$1280 resolution) using a search threshold of $\alpha=0.03$.
}
\label{tab:hunyuan720p_masks}
\begin{tabular}{ccccccrr}
\toprule
\multicolumn{2}{c}{\textbf{Mask Configuration}} & \multicolumn{2}{c}{\textbf{Head Count}} \\
\cmidrule(lr){1-2} \cmidrule(lr){3-4}
{$(T_w, T_o, T_s)$} & {$(S_w, S_o, S_s)$} & {\# Heads} & {\% Total} \\
\midrule
\multicolumn{2}{c}{full window} & 14400 & 20.00 \\
(11, 0, $\infty$) & (1, 0, 1) & 6223  & 8.64  \\
(1, 0, $\infty$)  & (5, 0, 6) & 4937  & 6.86  \\
(3, 2, $\infty$)  & (5, 0, 6) & 3995  & 5.55  \\
(11, 0, $\infty$) & (3, 0, 1) & 3964  & 5.51  \\
(11, 0, $\infty$) & (1, 0, 3) & 3511  & 4.88  \\
(11, 0, $\infty$) & (3, 0, 3) & 2898  & 4.03  \\
(1, 2, 2)  & (5, 0, 6) & 2385  & 3.31  \\
(3, 1, $\infty$)  & (5, 0, 6) & 2348  & 3.26  \\
(3, 0, $\infty$)  & (5, 0, 6) & 2090  & 2.90  \\
\midrule
\multicolumn{2}{l}{All Other Configurations} & 25249 & 35.06 \\
\midrule
\multicolumn{2}{l}{\textbf{Total}} & \bfseries 72000 & \bfseries 100.00 \\
\bottomrule
\end{tabular}
\end{table}

\section{Locality-Induced Periodic Diagonal Stripes}
\label{sec:locality_stripes}

In \cref{sec:unified_pattern_para}, we stated that video spatio-temporal locality alone can produce periodic diagonal stripes in the 2D attention heatmap. We now formalize this claim.

\noindent
\textbf{Proposition 1} (Locality-induced diagonal stripes).
Under the row-major token ordering $i = t \cdot HW + h \cdot W + w$, the locality conditions in \cref{eq:locality} produce periodic diagonal bands in the 2D attention heatmap:
\begin{enumerate}
\setlength{\itemsep}{1pt}
\setlength{\parsep}{0pt}
\setlength{\parskip}{0pt}
    \item \textbf{Temporal locality} ($|\Delta t| < \varepsilon_t$): For tokens at the same spatial position $(h,w)$, the 1D index difference is $\Delta i = \Delta t \cdot HW$, forming diagonal bands with stride $HW$ and half-width $\varepsilon_t \cdot HW$.
    \item \textbf{Column-wise locality} ($|\Delta h| < \varepsilon_h$): For tokens in the same column within a frame, $\Delta i = \Delta h \cdot W$, forming diagonal bands with stride $W$ and half-width $\varepsilon_h \cdot W$.
    \item \textbf{Row-wise locality} ($|\Delta w| < \varepsilon_w$): For tokens in the same row within a frame, $\Delta i = \Delta w$, forming a continuous band of width $2\varepsilon_w$ along the main diagonal.
\end{enumerate}
The composition of these conditions yields periodic diagonal stripes with stride determined by the token layout $(HW, W)$ and width determined by the locality radii $(\varepsilon_t, \varepsilon_h, \varepsilon_w)$.

\noindent
\textbf{Discussion.}
This proposition shows that locality alone suffices to produce periodic diagonal stripes without invoking RoPE periodicity: the stride is determined by the token layout ($HW$ for inter-frame, $W$ for intra-frame), and the width by the locality radii. For the 9.70\% of heads without a dominant RoPE channel (see \cref{sec:massive_m}), stripes arise purely from this mechanism. For heads with a dominant channel, RoPE periodicity further modulates the stride to $\mathcal{T}_\lambda$ (\cref{eq:period}) and introduces nonzero offset via $\phi^{(m^*_\lambda)}$, as described in the Joint effect paragraph of \cref{sec:unified_pattern_para}.

\section{Empirical Support for $m^*_\lambda$ in RoPE-3D}
\label{sec:massive_m}

% To further support the channel-dominance assumption in \cref{sec:unified_pattern_para} that for heads exhibiting structured
% sparse patterns, there exists a dominant channel

% there exists a single dominant channel $m^*_\lambda \in M_\lambda$  satisfying \cref{eq:dominance},

\begin{figure}[htbp]
    \centering
    \includegraphics[width=1\linewidth]{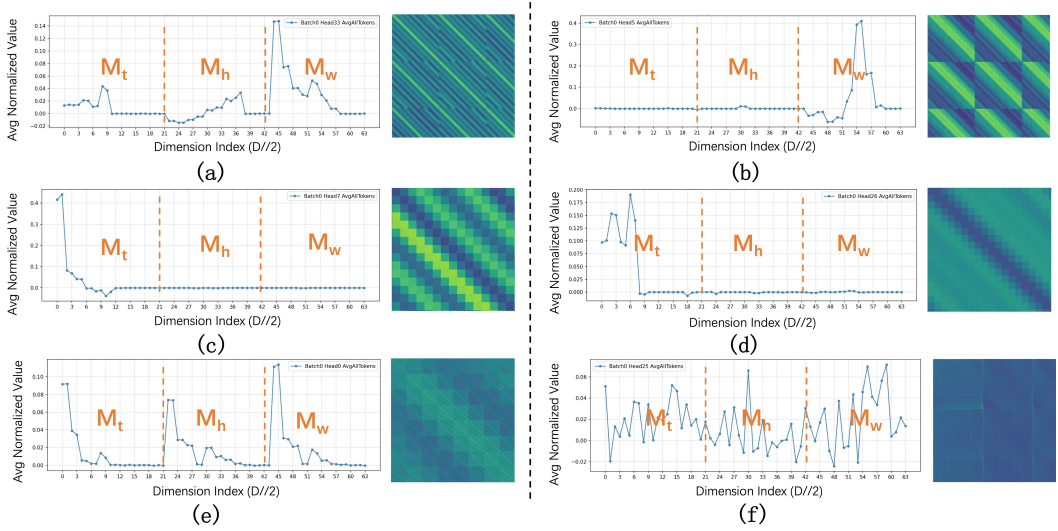}
    \caption{Empirical support for $m^*_\lambda$ in RoPE-3D}
    \label{fig:massive_m}
\end{figure}

To empirically validate the existence of a predominant channel $m^*_\lambda \in M_\lambda$ that satisfies \cref{eq:dominance} for structured attention patterns, we visualize the RoPE-channel contribution distributions of attention across a large number of attention heads in Wan~2.1 and HunyuanVideo, as shown in \cref{fig:massive_m}.
For each attention head, the RoPE-channel weights are computed by performing a dot product between the query and key vectors along each row, normalizing the resulting weights, and subsequently averaging over all rows at each corresponding position. The observations are summarized as follows:

1. Figures~(a)--(b) illustrate the RoPE-channel weight distributions associated with the intra-frame pattern. In this case, the dominant channel $m^*$ is consistently localized within $M_w$. Moreover, as the channel index within the $M_w$ dimension increases, 
the spacing between adjacent intra-frame diagonal stripes increases accordingly,
which corroborates the periodicity formulation in \cref{eq:period}.

2. Figures~(c)--(d) illustrate the RoPE-channel weight distributions associated with the inter-frame pattern. Here, the dominant channel $m^*$ is consistently localized within $M_t$. Analogously, as the channel index within the $M_t$ dimension increases, the spacing between adjacent inter-frame diagonal stripes widens accordingly,
consistent with the behavior described in \cref{eq:period}.

3. Figure~(e) corresponds to the hybrid pattern. Under this regime, pronounced dominant channels $m^*$ simultaneously emerge across multiple sub-groups $M_t$, $M_h$, and $M_w$, reflecting a superposition of multiple distinct attention patterns.

4. Figure~(f) corresponds to the uniform pattern. Under this regime, the weight distributions across all channels exhibit a disordered and non-structured configuration, with no discernible dominant channel.

Quantitatively, we measure the dominance ratio as $\max_m w_m / \mathrm{median}_m w_m$ for each head, where $w_m$ denotes the normalized RoPE-channel weight. On Wan~2.1, \textbf{87.55\%} of all attention heads exhibit a dominance ratio exceeding $3\times$, confirming that the single-channel dominance assumption underlying \cref{eq:dominance} holds for the vast majority of heads. Through experimental observation, among the remaining 12.45\% of heads, \textbf{2.74\%} correspond to the uniform pattern, while the remaining \textbf{9.70\%} still exhibit diagonal stripe structures. These stripes arise from the spatio-temporal locality ($\varepsilon_t$, $\varepsilon_h$, $\varepsilon_w$) described in \cref{eq:locality}, and can likewise be efficiently sparsified by \texttt{PSA}.
As formalized in \cref{sec:locality_stripes}, locality alone produces periodic diagonal stripes with stride determined by the token layout ($HW$, $W$), confirming that these heads are not exceptions but a natural sub-case of the unified stripe structure.
To verify that PSA effectively handles these locality-driven heads, we separately evaluate the 9.70\% of heads without a dominant RoPE channel. At 60\% sparsity, these heads achieve an average cosine similarity of \textbf{0.9872} with full attention, confirming that the locality-driven stripes are equally well captured by $\mathcal{M}$.

% Note that although the uniform pattern does not exhibit an explicit dominant $m^*$, it can be interpreted as a special case where the locality radii ($\varepsilon_t$, $\varepsilon_h$, $\varepsilon_w$) in \cref{eq:locality} are sufficiently large that the resulting diagonal stripes span the entire heatmap, effectively producing a uniform attention distribution. 
% Note that although the uniform pattern does not exhibit an explicit dominant $m^*$, it can be interpreted as a special case where the locality radii ($\varepsilon_t$, $\varepsilon_h$, $\varepsilon_w$) in \cref{eq:locality} are sufficiently large that the resulting diagonal stripes span the entire heatmap, effectively producing a uniform attention distribution. In practice, \texttt{PSA} can handle this pattern through either a wide diagonal-stripe mask or a uniform checkerboard-style downsampled mask (as illustrated in \cref{fig:param_vis}~(c)). \texttt{PSA-Search} automatically selects the most suitable mask---the one achieving the highest sparsity under the error threshold $\alpha$.

Although the uniform pattern lacks a dominant $m^*$, PSA-Search 
effectively handles these heads: the discovered masks fall into 
two categories—wide-stripe configurations (large $T_w$ or $S_w$) 
and checkerboard patterns (periodic small-stride masks)—both 
of which achieve faithful approximations of the uniform 
distribution (cosine similarity $>$ 0.98) while maintaining 
meaningful sparsity. This confirms that the PSA mask space 
is expressive enough to cover all four observed pattern types.

\section{Heatmaps of Different Heads in CogVideoX}
\label{sec:heat_map_cogvideo}

\begin{figure}[t]
    \centering
    \includegraphics[width=0.55\linewidth]{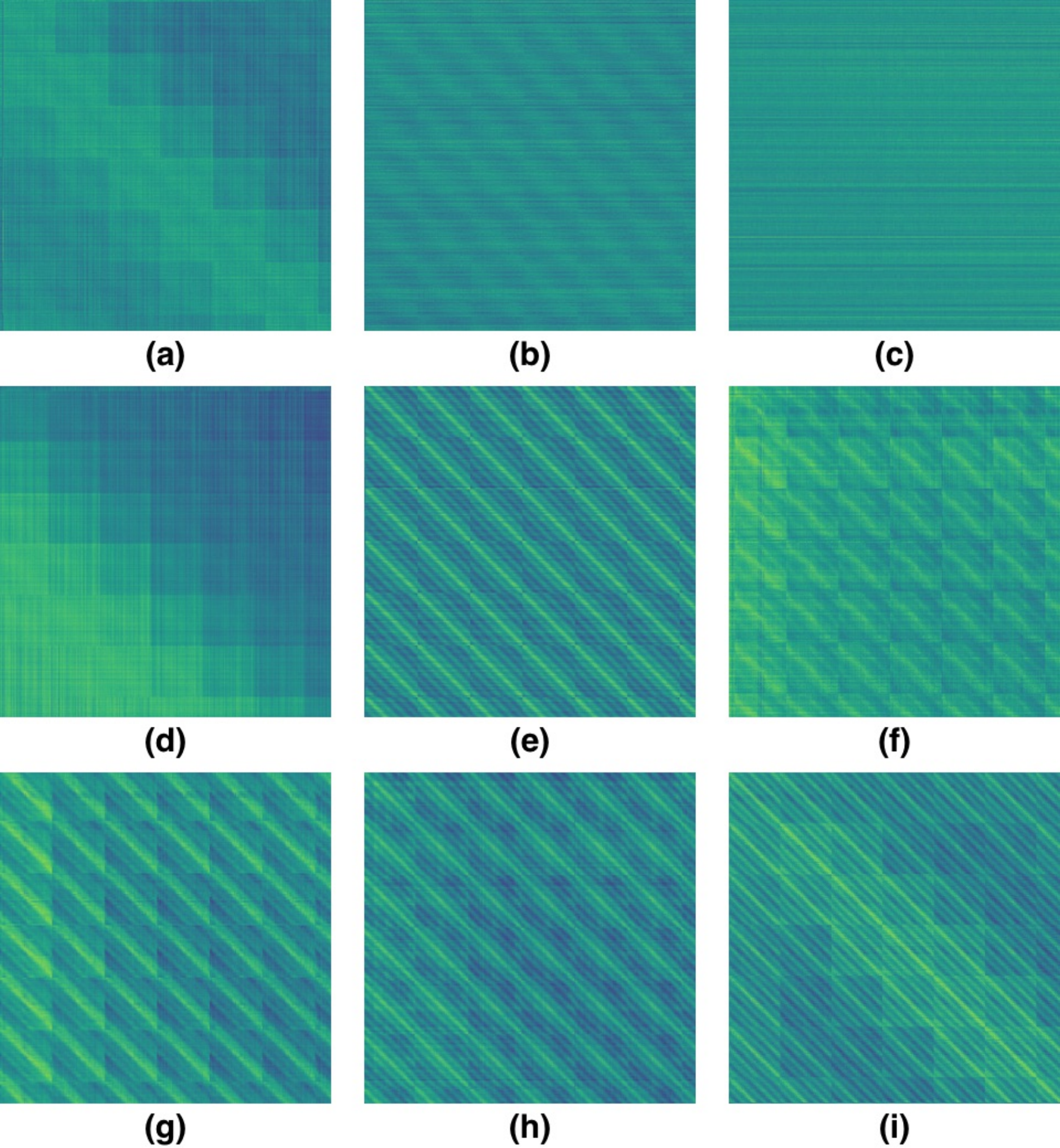}
    \caption{CogVideoX Heatmap Patterns.}
    \label{fig:cogvideo-heatmap}
\end{figure}

As shown in \cref{fig:cogvideo-heatmap}, we analyze various patterns in the attention heatmaps of CogVideoX, which exhibit behaviors consistent with HunyuanVideo and Wan~2.1. Four distinct pattern types are identified:

\begin{itemize}
    \item Figures~(b), (e)-(h) demonstrate \textbf{intra-frame attention patterns}. These patterns are characterized by diagonal structures with varying widths, strides, and offsets within each frame, with the pattern repeating identically across all frames.
    
    \item Figures~(a), (d) showcase \textbf{inter-frame attention patterns}. In these patterns, attention weights within each frame are relatively uniform, while diagonal band structures with varying widths, strides, and offsets emerge between frames.
    
    \item Figure~(i) presents a \textbf{hybrid pattern}, which can be interpreted as a combination of intra-frame and inter-frame attention patterns.
    
    \item Figure~(c) displays \textbf{uniform pattern}, where attention is uniformly distributed across all spatio-temporal positions.
\end{itemize}

Through further examination of the CogVideoX heatmap, we observe that it demonstrates sparse patterns analogous to those exhibited by Wan 2.1 and HunyuanVideo. The underlying reason is that these models all utilize DiT architectures for video generation, maintain structural similarities, and optimize toward identical objectives. Consequently, their weight distributions should also converge to comparable configurations, suggesting that the periodic locality characteristic of attention maps is universally applicable across DiT-based video generation scenarios. This demonstrates the general adaptability of \texttt{PSA}.

\section{Heatmap Visualization Across Different Prompts}
\label{sec:heat_map_diff_prompt}

\begin{figure*}
    \centering
    \includegraphics[width=1\linewidth]{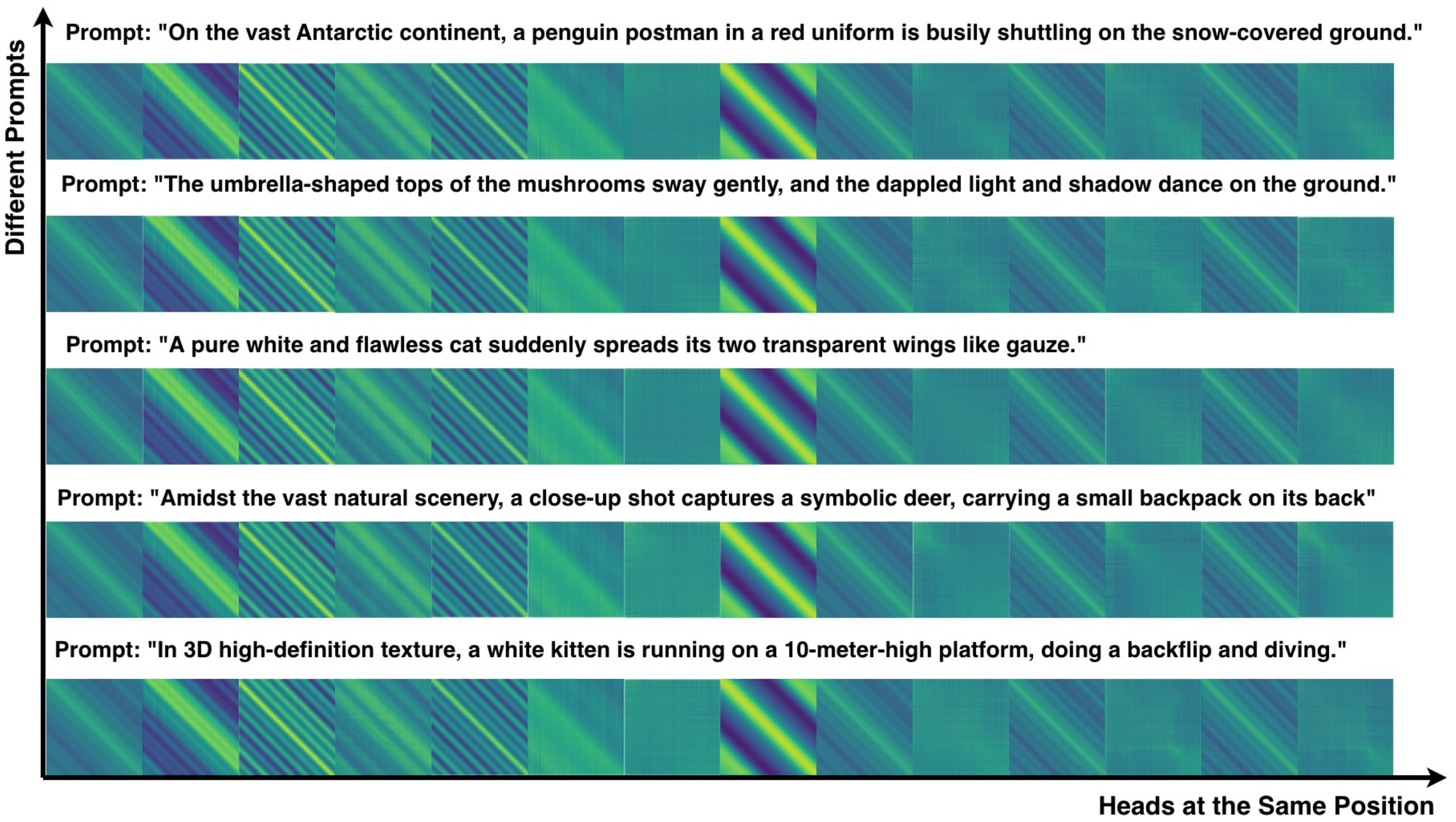}
    \caption{Heatmap Comparison Between Prompts.}
    \label{fig:why_heat_map}
\end{figure*}

As shown in \cref{fig:why_heat_map}, it can be intuitively observed that the heatmaps of the same head across different prompts exhibit extremely high similarity (cosine similarity $>$ 0.99).

\section{Loss Threshold Sensitivity Analysis}
\label{sec:Ablation_alpha}

\begin{figure}[t]
    \centering
    \begin{subfigure}[b]{0.32\textwidth}
        \centering
        \includegraphics[width=\linewidth]{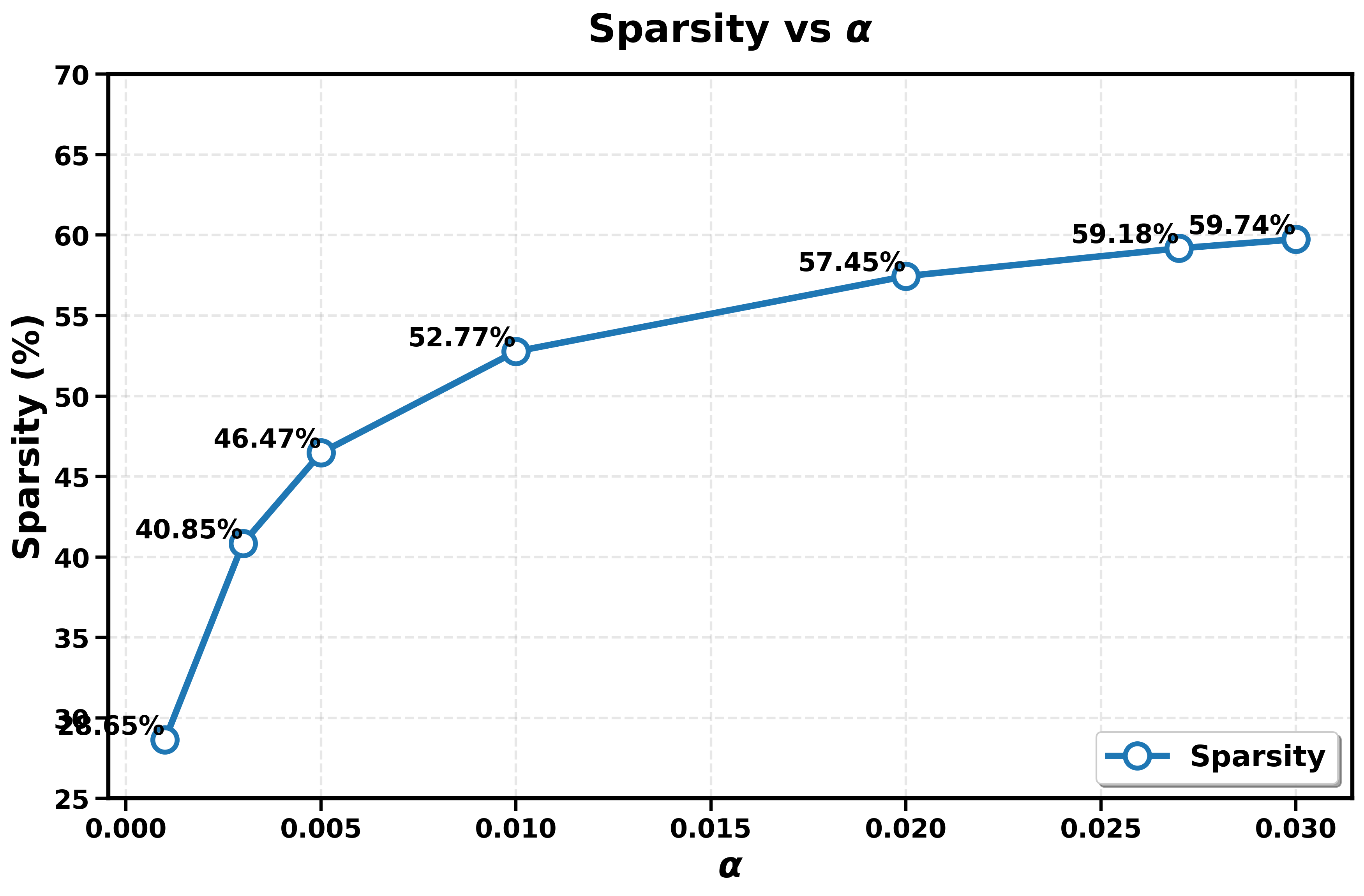}
        \caption{HunyuanVideo ($720\times1280$)}
        \label{fig:alpha-hy720}
    \end{subfigure}
    \hfill
    \begin{subfigure}[b]{0.32\textwidth}
        \centering
        \includegraphics[width=\linewidth]{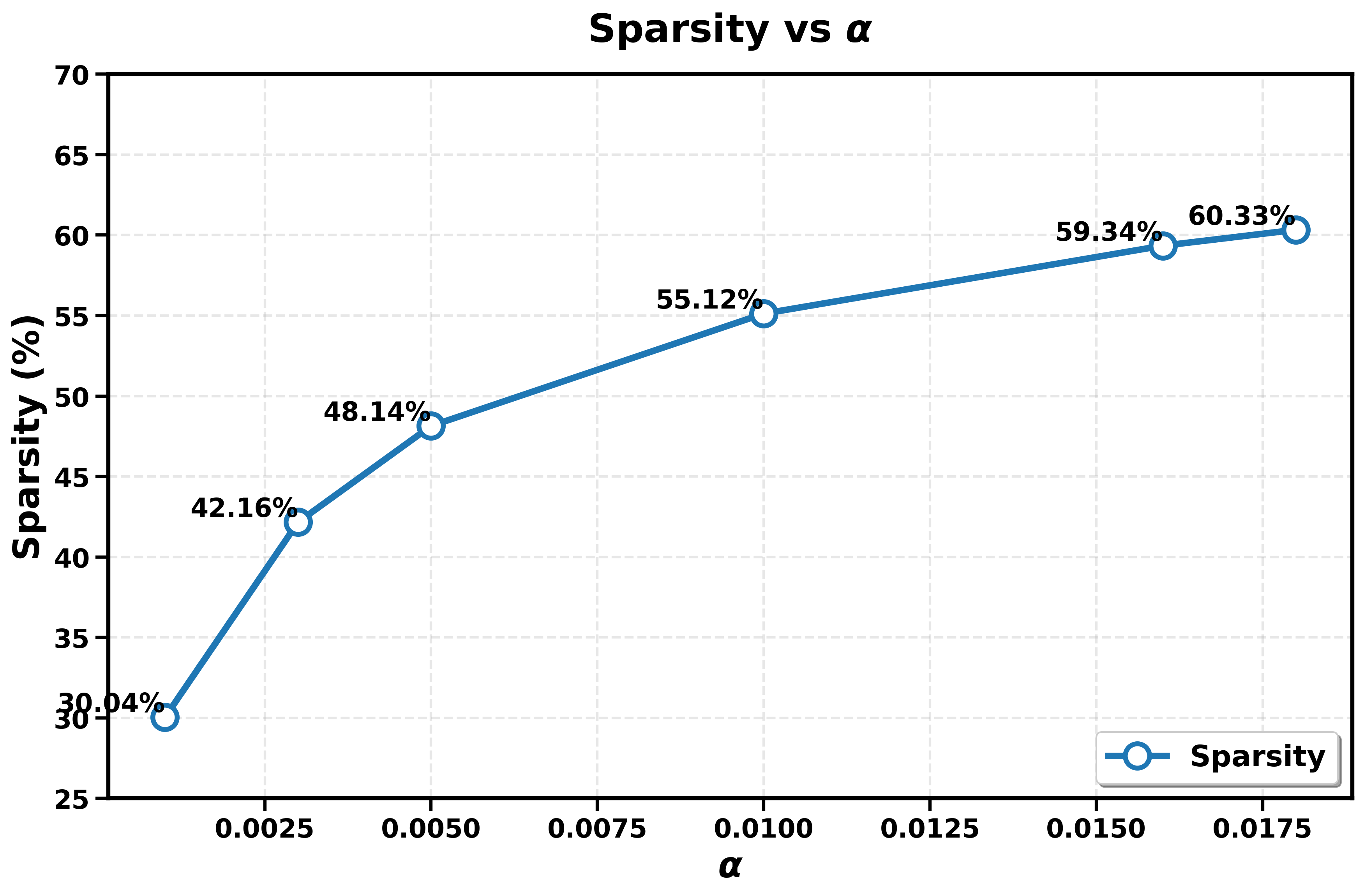}
        \caption{HunyuanVideo ($768\times1280$)}
        \label{fig:alpha-hy768}
    \end{subfigure}
    \hfill
    \begin{subfigure}[b]{0.32\textwidth}
        \centering
        \includegraphics[width=\linewidth]{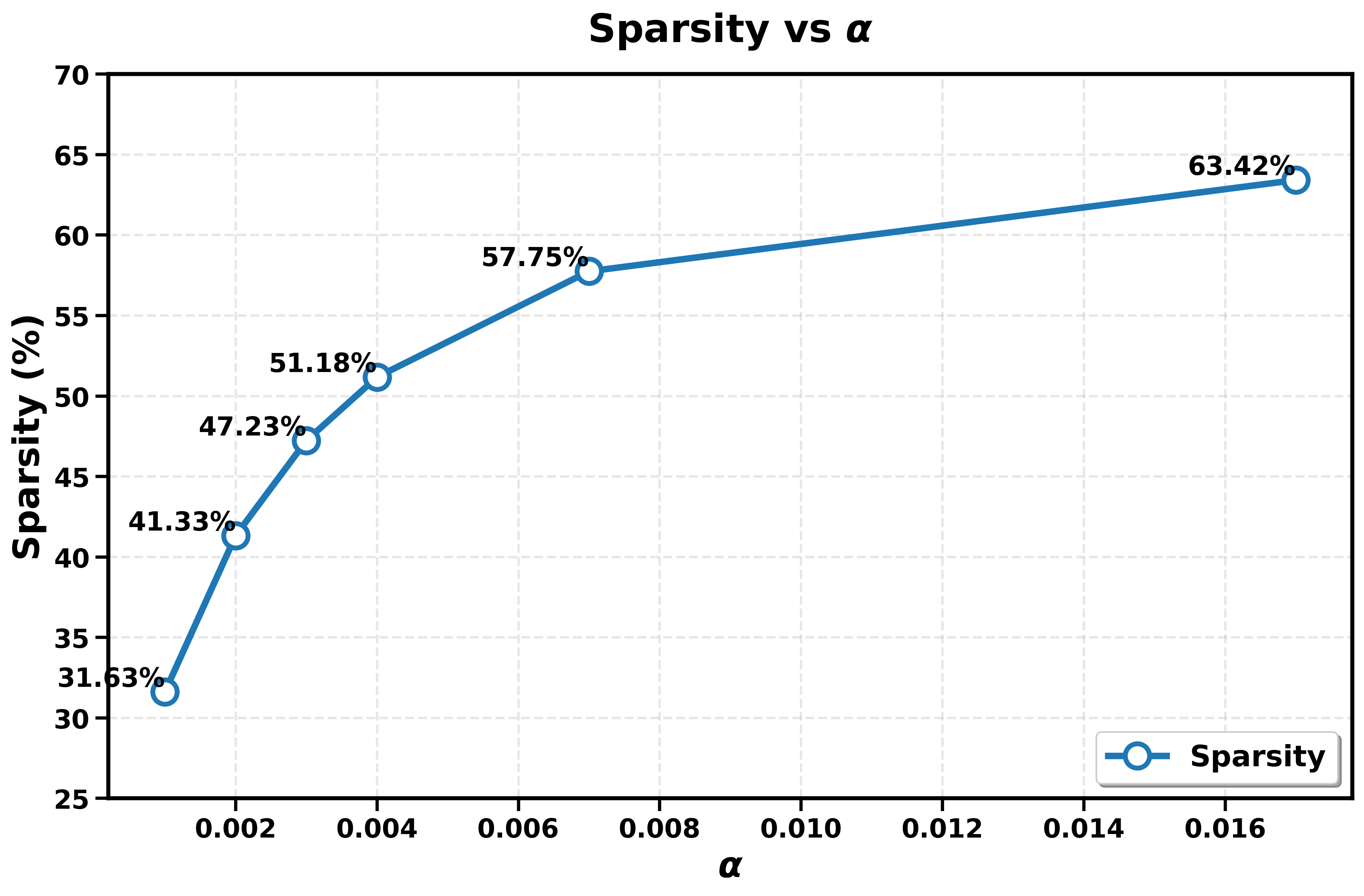}
        \caption{Wan~2.1 14B}
        \label{fig:alpha-wan}
    \end{subfigure}
    \caption{Sparsity variation with different $\alpha$ values across different models.}
    \label{fig:alpha-ablation}
\end{figure}

As illustrated in \cref{fig:alpha-ablation}, we perform loss threshold sensitivity analysis on various $\alpha$ values for HunyuanVideo and Wan~2.1. The experimental results indicate that larger $\alpha$ values correspond to higher tolerable loss thresholds, which given the design principle of \texttt{PSA-Search}, result in increased sparsity. To maintain sparsity at approximately 60\%, we configured $\alpha$ as 0.03, 0.018, and 0.007 for HunyuanVideo $(720\times1280)$, HunyuanVideo $(768\times1280)$, and Wan~2.1, respectively. This observation reveals that Wan~2.1 exhibits higher sensitivity to loss, thereby necessitating a smaller $\alpha$ value to attain equivalent sparsity levels.

% \begin{figure}[t]
%     \centering
%     \includegraphics[width=0.5\linewidth]{pic/sec5/alpha_ablation_hy720.png}
%     \caption{Sparsity variation with different $\alpha$ values on HunyuanVideo ($720\times1280$) }
%     \label{fig:alpha-hy720}
% \end{figure}

% \begin{figure}[t]
%     \centering
%     \includegraphics[width=0.5\linewidth]{pic/sec5/alpha_ablation_hy768.png}
%     \caption{Sparsity variation with different $\alpha$ values on HunyuanVideo ($768\times1280$) }
%     \label{fig:alpha-hy768}
% \end{figure}

% \begin{figure}[t]
%     \centering
%     \includegraphics[width=0.5\linewidth]{pic/sec5/alpha_ablation_wan.png}
%     \caption{Sparsity variation with different $\alpha$ values on Wan 2.1 }
%     \label{fig:alpha-wan}
% \end{figure}

Furthermore, we obtain $\alpha$ values at various sparsity levels through our adaptive search algorithm for reference. As shown in \cref{tab:alpha-sparsity-all}, setting the corresponding alpha value for a given sparsity level ensures optimal acceleration while minimizing the loss. Note that changing $\alpha$ does not require a time-consuming re-search, as \texttt{PSA-Search} saves the \texttt{HeadResults} for each head; consequently, after each adjustment of $\alpha$, only the re-execution of \cref{alg:psa-Search_concise} (lines 10–21) is needed, taking approximately 2 minutes.

% \begin{table}[htbp]
%     \centering
%     \caption{Alpha-sparsity correspondence for HunyuanVideo ($720\times1280$).}
%     \label{tab:alpha-sparsity-hy720}
%     \begin{tabular}{cc}
%         \toprule
%         Sparsity (\%) & $\alpha$ \\
%         \midrule
%         18.72 & 0.0003 \\
%         28.65 & 0.001  \\
%         40.85 & 0.003  \\
%         49.73 & 0.007  \\
%         62.80 & 0.06   \\
%         67.22 & 0.6    \\
%         \bottomrule
%     \end{tabular}
% \end{table}

% \begin{table}[htbp]
%     \centering
%     \caption{Alpha-sparsity correspondence for HunyuanVideo ($768\times1280$).}
%     \label{tab:alpha_sparsity}
%     \begin{tabular}{cc}
%         \toprule
%         Sparsity (\%) & $\alpha$ \\
%         \midrule
%         20.59 & 0.0003 \\
%         30.04 & 0.0010  \\
%         42.16 & 0.0030  \\
%         51.66 & 0.0070  \\
%         60.33 & 0.0180  \\
%         69.09 & 0.0600   \\
%         \bottomrule
%     \end{tabular}
%     \label{tab:alpha-sparsity-hy768}
% \end{table}

% \begin{table}[htbp]
%     \centering
%     \caption{Alpha-sparsity correspondence for Wan~2.1.}
%     \begin{tabular}{cc}
%         \toprule
%         Sparsity (\%) & $\alpha$ \\
%         \midrule
%         19.65 & 0.0003 \\
%         31.63 & 0.0010  \\
%         42.05 & 0.0021 \\
%         51.51 & 0.0041 \\
%         61.22 & 0.0110  \\
%         65.46 & 0.3000    \\
%         \bottomrule
%     \end{tabular}
%     \label{tab:alpha-sparsity-wan21}
% \end{table}

\begin{table}[htbp]
    \centering
    \caption{Alpha-sparsity correspondence for different models.}
    \label{tab:alpha-sparsity-all}
    \begin{tabular}{cc@{\hskip 2em}cc@{\hskip 2em}cc}
        \toprule
        \multicolumn{2}{c}{HunyuanVideo ($720\times1280$)} & \multicolumn{2}{c}{HunyuanVideo ($768\times1280$)} & \multicolumn{2}{c}{Wan~2.1 14B} \\
        \cmidrule(lr){1-2} \cmidrule(lr){3-4} \cmidrule(lr){5-6}
        Sparsity (\%) & $\alpha$ & Sparsity (\%) & $\alpha$ & Sparsity (\%) & $\alpha$ \\
        \midrule
        18.72 & 0.0003  & 20.59 & 0.0003 & 19.65 & 0.0003 \\
        28.65 & 0.0010  & 30.04 & 0.0010 & 31.63 & 0.0010 \\
        40.85 & 0.0030  & 42.16 & 0.0030 & 42.05 & 0.0021 \\
        49.73 & 0.0070  & 51.66 & 0.0070 & 51.51 & 0.0041 \\
        62.80 & 0.0600  & 60.33 & 0.0180 & 61.22 & 0.0110 \\
        67.22 & 0.6000  & 69.09 & 0.0600 & 65.46 & 0.3000  \\
        \bottomrule
    \end{tabular}
\end{table}

\section{End-to-End Performance and Quality Across Varying Sparsity Levels}
\label{sec:e2e-comp-diff-sparsity}
Beyond our initial analysis at 60\% sparsity to match STA's fixed sparsity (58.37\%) for a direct comparison, we evaluate 10–70\% sparsity (\cref{tab:e2e_robustness}). Our method consistently outperforms SVG2 in speed/quality, and achieves a better speed–quality trade-off than SVG.
SVG only achieves marginal speedup at 70\% sparsity while suffering severe quality degradation.

\begin{table}[h!]
\centering
\caption{End-to-end performance and quality comparison across varying sparsity on HunyuanVideo (768 $\times$ 1280).}
\label{tab:e2e_robustness}
\footnotesize
\renewcommand{\arraystretch}{0.95} % Slightly reduce row height for compactness
\begin{tabular}{lcccccc}
\toprule
\textbf{Method} & \textbf{Sparsity} & \textbf{E2E(s)} & \textbf{PSNR$\uparrow$} & \textbf{SSIM$\uparrow$} & \textbf{LPIPS$\downarrow$} \\
\midrule
% FA3 Baseline & 0\% & 1106.36 & - & - & - & - \\
% \midrule
\multirow{4}{*}{SVG} 
& 10.48\% & 1393.25  & 30.34 & 0.9144 & 0.1247 \\
& 30.39\% & 1116.87  & 27.68 & 0.8862 & 0.1552 \\
& 50.32\% & 865.78 & 25.67 & 0.8494 & 0.1914 \\
& 70.16\% & \textbf{622.37} & 21.21 & 0.7771 & 0.2651 \\
\midrule
\multirow{4}{*}{SVG2} 
& 10.49\% & 1333.93  & 31.87 & 0.9076 & 0.1206 \\
& 30.72\% & 1129.91 & 26.75 & 0.7753 & 0.2040 \\
& 50.67\% & 945.26  & 26.18 & 0.7314 & 0.2352 \\
& 70.93\% & 694.07  & 24.54 & 0.6610 & 0.2985 \\
\midrule
\multirow{4}{*}{\textbf{Ours}} 
& \textbf{11.28\%} & \textbf{1057.19}  & \textbf{38.83} & \textbf{0.9647} & \textbf{0.0526} \\
& \textbf{30.04\%} & \textbf{941.44}  & \textbf{33.99} & \textbf{0.9318} & \textbf{0.0812} \\
& \textbf{51.66\%} & \textbf{776.87}  & \textbf{30.21} & \textbf{0.8952} & \textbf{0.1244} \\
& \textbf{69.09\%} & 632.86  & \textbf{24.58} & \textbf{0.8220} & \textbf{0.2340} \\
\bottomrule
\end{tabular}
\end{table}

\section{Qualitative Comparison with Baselines}
\label{sec:qualitative}

\begin{figure*}
    \centering
    \includegraphics[width=1\linewidth]{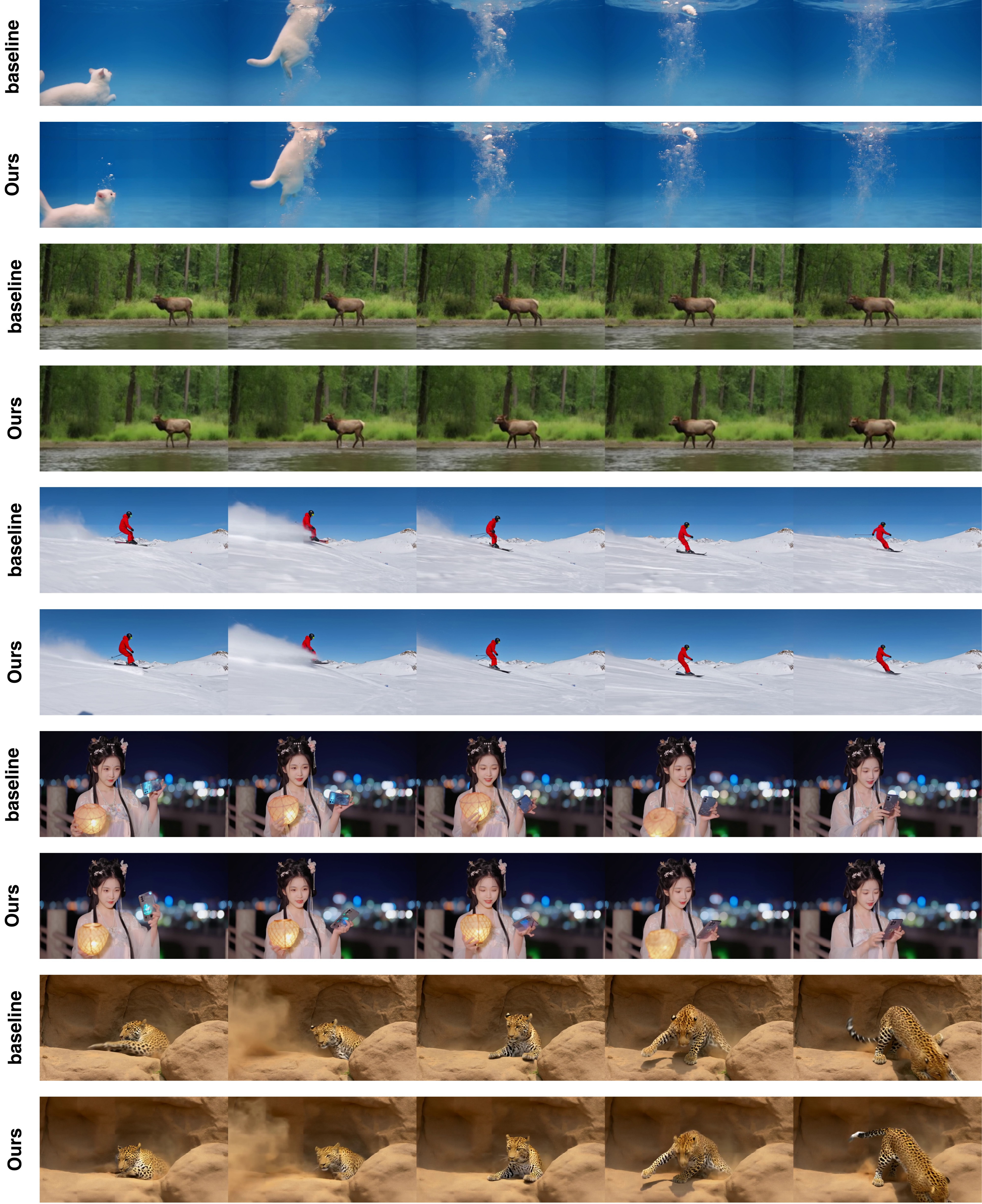}
    \caption{HunyuanVideo(720$\times$1280) FA3 Baseline vs. Ours.}
    \label{fig:hy720-pic}
\end{figure*}

\begin{figure*}
    \centering
    \includegraphics[width=1\linewidth]{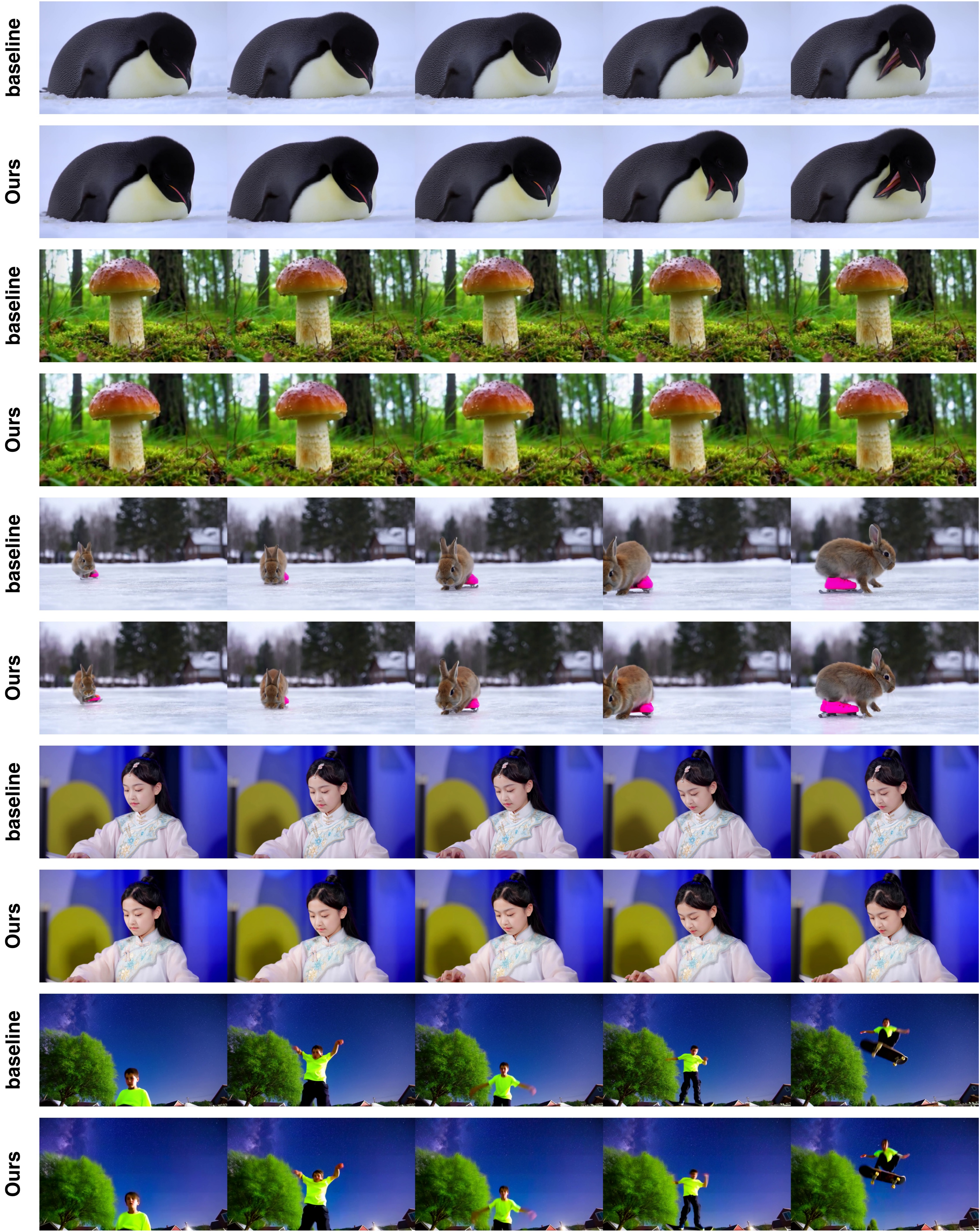}
    \caption{HunyuanVideo(768$\times$1280) FA3 Baseline vs. Ours.}
    \label{fig:hy768-pic}
\end{figure*}

\begin{figure*}
    \centering
    \includegraphics[width=1\linewidth]{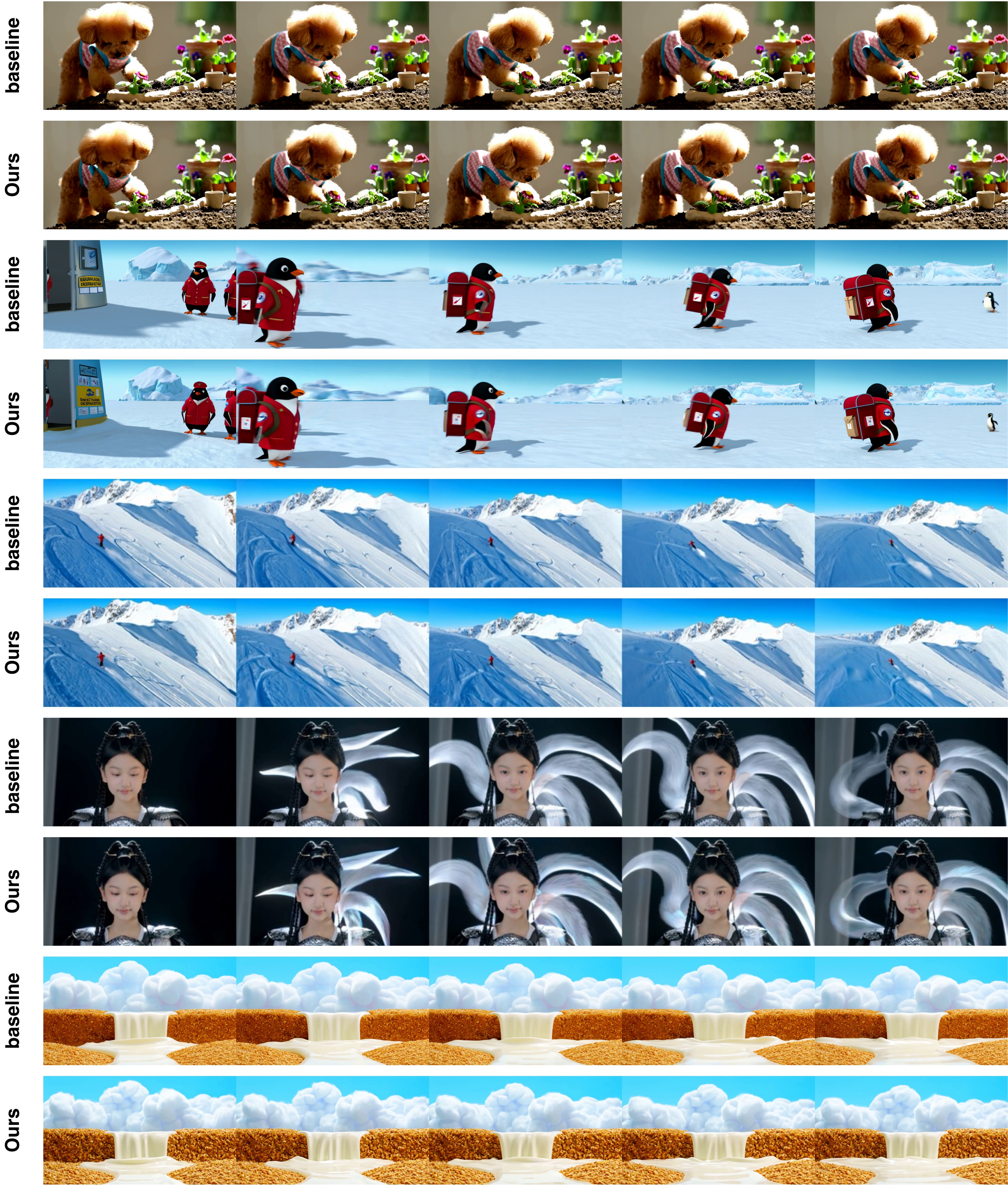}
    \caption{Wan~2.1(720$\times$1280) FA3 Baseline vs. Ours.}
    \label{fig:wan-pic}
\end{figure*}

As shown in \cref{fig:hy720-pic,fig:hy768-pic,fig:wan-pic}, we execute multiple prompts across several models. By comparing the extracted video frames with the baseline using FlashAttention-3, we can observe that the quality is nearly lossless.

%% file: checklist.tex
\section*{NeurIPS Paper Checklist}

\begin{enumerate}

\item {\bf Claims}
    \item[] Question: Do the main claims made in the abstract and introduction accurately reflect the paper's contributions and scope?
    \item[] Answer: \answerYes{}
    \item[] Justification: The abstract and introduction clearly state three contributions of PSA (parameterized representation, efficient kernel, automatic search), which are consistent with the theoretical and experimental results in Sections 3--4.

\item {\bf Limitations}
    \item[] Question: Does the paper discuss the limitations of the work performed by the authors?
    \item[] Answer: \answerYes{}
    \item[] Justification: The paper discusses limitations in Appendix~\ref{sec:limitations}.

\item {\bf Theory assumptions and proofs}
    \item[] Question: For each theoretical result, does the paper provide the full set of assumptions and a complete (and correct) proof?
    \item[] Answer: \answerYes{}
    \item[] Justification: The RoPE-3D attention logit decomposition (Eq.~1) and the dominant channel assumption (Eqs.~2--4) are presented with clear assumptions and derivations in Section~3.1.

\item {\bf Experimental result reproducibility}
    \item[] Question: Does the paper fully disclose all the information needed to reproduce the main experimental results of the paper to the extent that it affects the main claims and/or conclusions of the paper (regardless of whether the code and data are provided or not)?
    \item[] Answer: \answerYes{}
    \item[] Justification: The paper provides complete algorithm pseudocode (Algorithms 1--3), model configurations (Section~4.1), search parameter ranges (Appendix~B), and kernel implementation details (Appendix~A), sufficient to reproduce the main experiments.

\item {\bf Open access to data and code}
    \item[] Question: Does the paper provide open access to the data and code, with sufficient instructions to faithfully reproduce the main experimental results, as described in supplemental material?
    \item[] Answer: \answerYes{}
    \item[] Justification: The code is publicly available at \url{https://github.com/jxyjason/PSA}, with sufficient instructions to faithfully reproduce the main experimental results.

\item {\bf Experimental setting/details}
    \item[] Question: Does the paper specify all the training and test details (e.g., data splits, hyperparameters, how they were chosen, type of optimizer) necessary to understand the results?
    \item[] Answer: \answerYes{}
    \item[] Justification: Section~4.1 provides detailed descriptions of model configurations, evaluation metrics, baseline methods, search parameters, and experimental settings.

\item {\bf Experiment statistical significance}
    \item[] Question: Does the paper report error bars suitably and correctly defined or other appropriate information about the statistical significance of the experiments?
    \item[] Answer: \answerNo{}
    \item[] Justification: The randomness in video generation evaluation primarily stems from initial noise, while our comparison is between deterministic sparse attention and full attention outputs, so error bars are not required.

\item {\bf Experiments compute resources}
    \item[] Question: For each experiment, does the paper provide sufficient information on the computer resources (type of compute workers, memory, time of execution) needed to reproduce the experiments?
    \item[] Answer: \answerYes{}
    \item[] Justification: The paper states that all experiments are conducted on a single NVIDIA H100 GPU, and mentions in Section~3.3 that the search completes in 2.5 hours on 8 H100 GPUs.

\item {\bf Code of ethics}
    \item[] Question: Does the research conducted in the paper conform, in every respect, with the NeurIPS Code of Ethics \url{https://neurips.cc/public/EthicsGuidelines}?
    \item[] Answer: \answerYes{}
    \item[] Justification: This research is an efficiency optimization method for video generation and does not involve human subjects, sensitive data, or ethical risks.

\item {\bf Broader impacts}
    \item[] Question: Does the paper discuss both potential positive societal impacts and negative societal impacts of the work performed?
    \item[] Answer: \answerNA{}
    \item[] Justification: This research is a foundational efficiency optimization method that accelerates inference of existing models and is not directly tied to specific application deployments, with no direct path to negative societal impacts.

\item {\bf Safeguards}
    \item[] Question: Does the paper describe safeguards that have been put in place for responsible release of data or models that have a high risk for misuse (e.g., pre-trained language models, image generators, or scraped datasets)?
    \item[] Answer: \answerNA{}
    \item[] Justification: This paper does not release new pre-trained models or datasets; it only provides an attention efficiency optimization method.

\item {\bf Licenses for existing assets}
    \item[] Question: Are the creators or original owners of assets (e.g., code, data, models), used in the paper, properly credited and are the license and terms of use explicitly mentioned and properly respected?
    \item[] Answer: \answerYes{}
    \item[] Justification: All existing assets used in this paper are properly credited, and their licenses and terms of use are explicitly stated and fully respected.

\item {\bf New assets}
    \item[] Question: Are new assets introduced in the paper well documented and is the documentation provided alongside the assets?
    \item[] Answer: \answerYes{}
    \item[] Justification: The code is released at \url{https://github.com/jxyjason/PSA} with documentation provided in the repository; no other new assets are involved.

\item {\bf Crowdsourcing and research with human subjects}
    \item[] Question: For crowdsourcing experiments and research with human subjects, does the paper include the full text of instructions given to participants and screenshots, if applicable, as well as details about compensation (if any)?
    \item[] Answer: \answerNA{}
    \item[] Justification: This paper does not involve crowdsourcing or research with human subjects.

\item {\bf Institutional review board (IRB) approvals or equivalent for research with human subjects}
    \item[] Question: Does the paper describe potential risks incurred by study participants, whether such risks were disclosed to the subjects, and whether Institutional Review Board (IRB) approvals (or an equivalent approval/review based on the requirements of your country or institution) were obtained?
    \item[] Answer: \answerNA{}
    \item[] Justification: This paper does not involve research with human subjects.

\item {\bf Declaration of LLM usage}
    \item[] Question: Does the paper describe the usage of LLMs if it is an important, original, or non-standard component of the core methods in this research? Note that if the LLM is used only for writing, editing, or formatting purposes and does \emph{not} impact the core methodology, scientific rigor, or originality of the research, declaration is not required.
    \item[] Answer: \answerNA{}
    \item[] Justification: LLMs are not a component of the core method in this research; the core method is sparse attention optimization based on parameterized stripe attention.

\end{enumerate}

%% file: main.bib
@String(CVPR  = {IEEE Conf. Comput. Vis. Pattern Recog.})

@String(ICLR  = {Int. Conf. Learn. Represent.})

@String(CVPR  = {CVPR})

@String(ICLR  = {ICLR})

@inproceedings{peebles2023scalable,
  title={Scalable diffusion models with transformers},
  author={Peebles, William and Xie, Saining},
  booktitle={Proceedings of the IEEE/CVF international conference on computer vision},
  pages={4195--4205},
  year={2023}
}

@article{vaswani2017attention,
  title={Attention is all you need},
  author={Vaswani, Ashish and Shazeer, Noam and Parmar, Niki and Uszkoreit, Jakob and Jones, Llion and Gomez, Aidan N and Kaiser, {\L}ukasz and Polosukhin, Illia},
  journal={Advances in neural information processing systems},
  volume={30},
  year={2017}
}

@article{svg,
  title={Sparse VideoGen: Accelerating Video Diffusion Transformers with Spatial-Temporal Sparsity},
  author={Xi, Haocheng and Yang, Shuo and Zhao, Yilong and Xu, Chenfeng and Li, Muyang and Li, Xiuyu and Lin, Yujun and Cai, Han and Zhang, Jintao and Li, Dacheng and others},
  journal={arXiv preprint arXiv:2502.01776},
  year={2025}
}

@misc{spector2024thunderkittenssimplefastadorable,
      title={ThunderKittens: Simple, Fast, and Adorable AI Kernels}, 
      author={Benjamin F. Spector and Simran Arora and Aaryan Singhal and Daniel Y. Fu and Christopher Ré},
      year={2024},
      eprint={2410.20399},
      archivePrefix={arXiv},
      primaryClass={cs.LG},
      url={https://arxiv.org/abs/2410.20399}, 
}

@article{sta,
  title={Fast video generation with sliding tile attention},
  author={Zhang, Peiyuan and Chen, Yongqi and Su, Runlong and Ding, Hangliang and Stoica, Ion and Liu, Zhengzhong and Zhang, Hao},
  journal={arXiv preprint arXiv:2502.04507},
  year={2025}
}

@article{sparse_vdit,
  title={Sparse-vDiT: Unleashing the Power of Sparse Attention to Accelerate Video Diffusion Transformers},
  author={Chen, Pengtao and Zeng, Xianfang and Zhao, Maosen and Ye, Peng and Shen, Mingzhu and Cheng, Wei and Yu, Gang and Chen, Tao},
  journal={arXiv preprint arXiv:2506.03065},
  year={2025}
}

@article{wu2025vmoba,
  title={VMoBA: Mixture-of-Block Attention for Video Diffusion Models},
  author={Wu, Jianzong and Hou, Liang and Yang, Haotian and Tao, Xin and Tian, Ye and Wan, Pengfei and Zhang, Di and Tong, Yunhai},
  journal={arXiv preprint arXiv:2506.23858},
  year={2025}
}

@article{svg2,
  title={Sparse VideoGen2: Accelerate Video Generation with Sparse Attention via Semantic-Aware Permutation},
  author={Yang, Shuo and Xi, Haocheng and Zhao, Yilong and Li, Muyang and Zhang, Jintao and Cai, Han and Lin, Yujun and Li, Xiuyu and Xu, Chenfeng and Peng, Kelly and others},
  journal={arXiv preprint arXiv:2505.18875},
  year={2025}
}

@article{sun2025vorta,
  title={VORTA: Efficient Video Diffusion via Routing Sparse Attention},
  author={Sun, Wenhao and Tu, Rong-Cheng and Ding, Yifu and Jin, Zhao and Liao, Jingyi and Liu, Shunyu and Tao, Dacheng},
  journal={arXiv preprint arXiv:2505.18809},
  year={2025}
}

@article{lu2022dpm,
  title={Dpm-solver: A fast ode solver for diffusion probabilistic model sampling in around 10 steps},
  author={Lu, Cheng and Zhou, Yuhao and Bao, Fan and Chen, Jianfei and Li, Chongxuan and Zhu, Jun},
  journal={Advances in neural information processing systems},
  volume={35},
  pages={5775--5787},
  year={2022}
}

@article{lu2025dpm++,
  title={Dpm-solver++: Fast solver for guided sampling of diffusion probabilistic models},
  author={Lu, Cheng and Zhou, Yuhao and Bao, Fan and Chen, Jianfei and Li, Chongxuan and Zhu, Jun},
  journal={Machine Intelligence Research},
  pages={1--22},
  year={2025},
  publisher={Springer}
}

@article{wan2025wan,
  title={Wan: Open and advanced large-scale video generative models},
  author={Wan, Team and Wang, Ang and Ai, Baole and Wen, Bin and Mao, Chaojie and Xie, Chen-Wei and Chen, Di and Yu, Feiwu and Zhao, Haiming and Yang, Jianxiao and others},
  journal={arXiv preprint arXiv:2503.20314},
  year={2025}
}

@article{kong2024hunyuanvideo,
  title={Hunyuanvideo: A systematic framework for large video generative models},
  author={Kong, Weijie and Tian, Qi and Zhang, Zijian and Min, Rox and Dai, Zuozhuo and Zhou, Jin and Xiong, Jiangfeng and Li, Xin and Wu, Bo and Zhang, Jianwei and others},
  journal={arXiv preprint arXiv:2412.03603},
  year={2024}
}

@article{ddim,
  title={Denoising diffusion implicit models},
  author={Song, Jiaming and Meng, Chenlin and Ermon, Stefano},
  journal={arXiv preprint arXiv:2010.02502},
  year={2020}
}

@inproceedings{esser2024rectifiedflow,
  title={Scaling rectified flow transformers for high-resolution image synthesis},
  author={Esser, Patrick and Kulal, Sumith and Blattmann, Andreas and Entezari, Rahim and M{\"u}ller, Jonas and Saini, Harry and Levi, Yam and Lorenz, Dominik and Sauer, Axel and Boesel, Frederic and others},
  booktitle={Forty-first international conference on machine learning},
  year={2024}
}

@article{lipman2022flowmatching,
  title={Flow matching for generative modeling},
  author={Lipman, Yaron and Chen, Ricky TQ and Ben-Hamu, Heli and Nickel, Maximilian and Le, Matt},
  journal={arXiv preprint arXiv:2210.02747},
  year={2022}
}

@inproceedings{hassani2023neighborhood,
  title={Neighborhood attention transformer},
  author={Hassani, Ali and Walton, Steven and Li, Jiachen and Li, Shen and Shi, Humphrey},
  booktitle={Proceedings of the IEEE/CVF conference on computer vision and pattern recognition},
  pages={6185--6194},
  year={2023}
}

@misc{jacobs2023deepspeedulyssesoptimizationsenabling,
      title={DeepSpeed Ulysses: System Optimizations for Enabling Training of Extreme Long Sequence Transformer Models}, 
      author={Sam Ade Jacobs and Masahiro Tanaka and Chengming Zhang and Minjia Zhang and Shuaiwen Leon Song and Samyam Rajbhandari and Yuxiong He},
      year={2023},
      eprint={2309.14509},
      archivePrefix={arXiv},
      primaryClass={cs.LG},
      url={https://arxiv.org/abs/2309.14509}, 
}

@article{lu2025moba,
  title={Moba: Mixture of block attention for long-context llms},
  author={Lu, Enzhe and Jiang, Zhejun and Liu, Jingyuan and Du, Yulun and Jiang, Tao and Hong, Chao and Liu, Shaowei and He, Weiran and Yuan, Enming and Wang, Yuzhi and others},
  journal={arXiv preprint arXiv:2502.13189},
  year={2025}
}

@inproceedings{dao2023flashattention2,
  title={Flash{A}ttention-2: Faster Attention with Better Parallelism and Work Partitioning},
  author={Dao, Tri},
  booktitle={International Conference on Learning Representations (ICLR)},
  year={2024}
}

@article{shah2024flashattention,
  title={Flashattention-3: Fast and accurate attention with asynchrony and low-precision},
  author={Shah, Jay and Bikshandi, Ganesh and Zhang, Ying and Thakkar, Vijay and Ramani, Pradeep and Dao, Tri},
  journal={Advances in Neural Information Processing Systems},
  volume={37},
  pages={68658--68685},
  year={2024}
}

@article{yang2024cogvideox,
  title={Cogvideox: Text-to-video diffusion models with an expert transformer},
  author={Yang, Zhuoyi and Teng, Jiayan and Zheng, Wendi and Ding, Ming and Huang, Shiyu and Xu, Jiazheng and Yang, Yuanming and Hong, Wenyi and Zhang, Xiaohan and Feng, Guanyu and others},
  journal={arXiv preprint arXiv:2408.06072},
  year={2024}
}

@inproceedings{zhang2018perceptualLPIPS,
  title={The Unreasonable Effectiveness of Deep Features as a Perceptual Metric},
  author={Zhang, Richard and Isola, Phillip and Efros, Alexei A and Shechtman, Eli and Wang, Oliver},
  booktitle={CVPR},
  year={2018}
}

@article{zhang2024evaluationagentVBench,
    title = {Evaluation Agent: Efficient and Promptable Evaluation Framework for Visual Generative Models},
    author = {Zhang, Fan and Tian, Shulin and Huang, Ziqi and Qiao, Yu and Liu, Ziwei},
    journal={arXiv preprint arXiv:2412.09645},
    year = {2024}
}

@misc{xia2025trainingfreeadaptivesparseattentionadaspa,
      title={Training-free and Adaptive Sparse Attention for Efficient Long Video Generation}, 
      author={Yifei Xia and Suhan Ling and Fangcheng Fu and Yujie Wang and Huixia Li and Xuefeng Xiao and Bin Cui},
      year={2025},
      eprint={2502.21079},
      archivePrefix={arXiv},
      primaryClass={cs.CV},
      url={https://arxiv.org/abs/2502.21079}, 
}

@misc{adnan2025foresightadaptivelayerreuse,
      title={Foresight: Adaptive Layer Reuse for Accelerated and High-Quality Text-to-Video Generation}, 
      author={Muhammad Adnan and Nithesh Kurella and Akhil Arunkumar and Prashant J. Nair},
      year={2025},
      eprint={2506.00329},
      archivePrefix={arXiv},
      primaryClass={cs.LG},
      url={https://arxiv.org/abs/2506.00329}, 
}

@misc{chen2025rainfusionadaptivevideogeneration,
      title={RainFusion: Adaptive Video Generation Acceleration via Multi-Dimensional Visual Redundancy}, 
      author={Aiyue Chen and Bin Dong and Jingru Li and Jing Lin and Kun Tian and Yiwu Yao and Gongyi Wang},
      year={2025},
      eprint={2505.21036},
      archivePrefix={arXiv},
      primaryClass={cs.CV},
      url={https://arxiv.org/abs/2505.21036}, 
}

@misc{zhang2025trainingfreeefficientvideogeneration,
      title={Training-Free Efficient Video Generation via Dynamic Token Carving}, 
      author={Yuechen Zhang and Jinbo Xing and Bin Xia and Shaoteng Liu and Bohao Peng and Xin Tao and Pengfei Wan and Eric Lo and Jiaya Jia},
      year={2025},
      eprint={2505.16864},
      archivePrefix={arXiv},
      primaryClass={cs.CV},
      url={https://arxiv.org/abs/2505.16864}, 
}

@misc{pan2021vared2videoadaptiveredundancy,
      title={VA-RED$^2$: Video Adaptive Redundancy Reduction}, 
      author={Bowen Pan and Rameswar Panda and Camilo Fosco and Chung-Ching Lin and Alex Andonian and Yue Meng and Kate Saenko and Aude Oliva and Rogerio Feris},
      year={2021},
      eprint={2102.07887},
      archivePrefix={arXiv},
      primaryClass={cs.CV},
      url={https://arxiv.org/abs/2102.07887}, 
}

@misc{wang2025retakereducingtemporalknowledge,
      title={ReTaKe: Reducing Temporal and Knowledge Redundancy for Long Video Understanding}, 
      author={Xiao Wang and Qingyi Si and Jianlong Wu and Shiyu Zhu and Li Cao and Liqiang Nie},
      year={2025},
      eprint={2412.20504},
      archivePrefix={arXiv},
      primaryClass={cs.CV},
      url={https://arxiv.org/abs/2412.20504}, 
}

@misc{ye2025flashinferefficientcustomizableattention,
      title={FlashInfer: Efficient and Customizable Attention Engine for LLM Inference Serving}, 
      author={Zihao Ye and Lequn Chen and Ruihang Lai and Wuwei Lin and Yineng Zhang and Stephanie Wang and Tianqi Chen and Baris Kasikci and Vinod Grover and Arvind Krishnamurthy and Luis Ceze},
      year={2025},
      eprint={2501.01005},
      archivePrefix={arXiv},
      primaryClass={cs.DC},
      url={https://arxiv.org/abs/2501.01005}, 
}

@misc{yang2026attentionpatternsexistunifying,
      title={Why Attention Patterns Exist: A Unifying Temporal Perspective Analysis}, 
      author={Qingyue Yang and Jie Wang and Xing Li and Yinqi Bai and Xialiang Tong and Huiling Zhen and Jianye Hao and Mingxuan Yuan and Bin Li},
      year={2026},
      eprint={2601.21709},
      archivePrefix={arXiv},
      primaryClass={cs.CL},
      url={https://arxiv.org/abs/2601.21709}, 
}

@article{su2024roformer,
  title={Roformer: Enhanced transformer with rotary position embedding},
  author={Su, Jianlin and Ahmed, Murtadha and Lu, Yu and Pan, Shengfeng and Bo, Wen and Liu, Yunfeng},
  journal={Neurocomputing},
  volume={568},
  pages={127063},
  year={2024},
  publisher={Elsevier}
}

@misc{flux2024,
    author={Black Forest Labs},
    title={FLUX},
    year={2024},
    howpublished={\url{https://github.com/black-forest-labs/flux}},
}

@misc{liu2026mixturedistributionsmattersdynamic,
      title={Mixture of Distributions Matters: Dynamic Sparse Attention for Efficient Video Diffusion Transformers}, 
      author={Yuxi Liu and Yipeng Hu and Zekun Zhang and Kunze Jiang and Kun Yuan},
      year={2026},
      eprint={2601.11641},
      archivePrefix={arXiv},
      primaryClass={cs.CV},
      url={https://arxiv.org/abs/2601.11641}, 
}

@misc{liu2026ropeslr3dropedrivensparselowrank,
      title={RoPeSLR: 3D RoPE-driven Sparse-LowRank Attention for Efficient Diffusion Transformers}, 
      author={Yuxi Liu and Zekun Zhang and Yixiang Cai and Renjia Deng and Yutong He and Kun Yuan},
      year={2026},
      eprint={2605.20659},
      archivePrefix={arXiv},
      primaryClass={cs.CV},
      url={https://arxiv.org/abs/2605.20659}, 
}

@article{opensora,
  title={Open-sora: Democratizing efficient video production for all},
  author={Zheng, Zangwei and Peng, Xiangyu and Yang, Tianji and Shen, Chenhui and Li, Shenggui and Liu, Hongxin and Zhou, Yukun and Li, Tianyi and You, Yang},
  journal={arXiv preprint arXiv:2412.20404},
  year={2024}
}

@misc{li2024hunyuandit,
      title={Hunyuan-DiT: A Powerful Multi-Resolution Diffusion Transformer with Fine-Grained Chinese Understanding}, 
      author={Zhimin Li and Jianwei Zhang and Qin Lin and Jiangfeng Xiong and Yanxin Long and Xinchi Deng and Yingfang Zhang and Xingchao Liu and Minbin Huang and Zedong Xiao and Dayou Chen and Jiajun He and Jiahao Li and Wenyue Li and Chen Zhang and Rongwei Quan and Jianxiang Lu and Jiabin Huang and Xiaoyan Yuan and Xiaoxiao Zheng and Yixuan Li and Jihong Zhang and Chao Zhang and Meng Chen and Jie Liu and Zheng Fang and Weiyan Wang and Jinbao Xue and Yangyu Tao and Jianchen Zhu and Kai Liu and Sihuan Lin and Yifan Sun and Yun Li and Dongdong Wang and Mingtao Chen and Zhichao Hu and Xiao Xiao and Yan Chen and Yuhong Liu and Wei Liu and Di Wang and Yong Yang and Jie Jiang and Qinglin Lu and others},
      year={2024},
      eprint={2405.08748},
      archivePrefix={arXiv},
      primaryClass={cs.CV}
}
